\documentclass{article} 
\usepackage{iclr2027_conference,times}

\usepackage{amsmath,amsfonts,bm}

\def\eqref#1{equation~\ref{#1}}

\def\1{\bm{1}}

\DeclareMathAlphabet{\mathsfit}{\encodingdefault}{\sfdefault}{m}{sl}
\SetMathAlphabet{\mathsfit}{bold}{\encodingdefault}{\sfdefault}{bx}{n}

\usepackage{hyperref}
\usepackage{cleveref}
\usepackage{booktabs}
\usepackage{makecell}
\usepackage{graphicx}
\usepackage{subcaption}
\usepackage{float}
\usepackage{multirow}
\usepackage{url}
\usepackage{subcaption}
\usepackage{xcolor}
\usepackage{pgfplots}
\pgfplotsset{compat=1.18}
\usepackage{tabularx}
\usepackage{wrapfig}
\usepackage{array}

\usepackage[most]{tcolorbox}

\definecolor{questionbg}{HTML}{EFF4FA}
\definecolor{questionframe}{HTML}{526D91}

\newtcolorbox{questionbox}{
    colback=questionbg,
    colframe=questionframe,
    boxrule=0.7pt,
    arc=2mm,
    left=7pt,
    right=7pt,
    top=5pt,
    bottom=5pt,
    before skip=8pt,
    after skip=8pt,
    fontupper=\small
}

\definecolor{takeawaybg}{HTML}{F0F7F3}
\definecolor{takeawayframe}{HTML}{507563}

\newtcolorbox{takeawaybox}{
    colback=takeawaybg,
    colframe=takeawayframe,
    boxrule=0.7pt,
    arc=2mm,
    left=7pt,
    right=7pt,
    top=5pt,
    bottom=5pt,
    before skip=8pt,
    after skip=8pt,
    fontupper=\small
}

\definecolor{dropred}{HTML}{C62828}
\definecolor{risegreen}{HTML}{008545}

\providecommand{\srcell}[2]{}
\providecommand{\srorig}[1]{}
\providecommand{\srdown}[2]{}
\providecommand{\srup}[2]{}

\renewcommand{\srcell}[2]{%
    \makebox[5.5em][l]{%
        \makebox[2.6em][r]{\ensuremath{#1}}%
        \makebox[2.9em][l]{\ensuremath{{}_{#2}}}%
    }%
}
\renewcommand{\srorig}[1]{%
    \srcell{#1}{}%
}
\renewcommand{\srdown}[2]{%
    \srcell{#1}{\textcolor{dropred}{\downarrow #2}}%
}
\renewcommand{\srup}[2]{%
    \srcell{#1}{\textcolor{risegreen}{\uparrow #2}}%
}
\definecolor{mydarkblue}{rgb}{0,0.08,0.45}
\definecolor{darkred}{rgb}{0.55,0,0}
\usepackage{hyperref}
\hypersetup{
    colorlinks=true,
    citecolor=mydarkblue, 
    linkcolor=darkred,    
    urlcolor=mydarkblue
}

\title{Encoded but Not in Control:\\ Revealing the Grounding Gap in Vision-Language Robot Policies}

\author{Shaohan Jiang$^{1,3}$\thanks{Equal contribution~ \texttt{shaohanjiang@link.cuhk.edu.cn}} \quad Jiahang Cao$^{1,2*}$ \quad Qiduo He$^{3*}$ \quad  Fengting Deng$^{3}$ \quad Kun Wu$^{4}$  \\ 
\textbf{Jingkai Sun$^{1}$ \quad Jiaxu Wang$^{5}$ \quad Qiang Zhang$^{6}$ \quad Qihao Zheng$^{2}$ \quad Chunfeng Song$^{2}$}\\
\textbf{Ping Luo$^{1}$ \quad Andrew F. Luo$^{1}$}\thanks{Corresponding author~ \texttt{aluo@hku.hk}}\\
$^1$HKU \quad $^2$Shanghai AI Lab \quad $^3$CUHK(SZ) \quad $^4$Horizon Robotics \quad $^5$CUHK \quad $^6$USTC \\
}

\iclrfinalcopy 
\begin{document}

\maketitle

\begin{abstract}
Instruction following is central to language-conditioned robot policies: language should determine what to do when the same scene permits multiple valid actions. Yet successful execution alone cannot establish whether a policy follows the instruction or infers the task from the scene. We study this ambiguity through \emph{scene-preserving instruction interventions}, using valid target substitutions, arbitrary nouns, and unrelated sentences while holding the scene fixed. We evaluate vision-language-action (VLA) policies and world-action models (WAMs) in simulation and in real-world experiments. Our analysis addresses three questions:
\textit{\textbf{(a)} Does task success imply instruction following?} When instructions request a different visible object, all evaluated policies predominantly approach and pick up the incorrect original target associated with the scene.
\textit{\textbf{(b)} Is this failure caused by language insensitivity?} Instruction perturbations affect task performance. A layerwise action lens shows intermediate action predictions respond to these perturbations. Linear probes accurately recover instructed targets, indicating modified instructions are encoded despite rarely determining target selection.
\textit{\textbf{(c)} Why does encoded language fail to control action?} Attention analysis indicates weak instruction-token contributions to action generation. Target-token attention can remain focused on the original object, revealing a mismatch between target encoding and visual grounding. UMAP and shared non-negative matrix factorization show target information remains accessible within representations increasingly organized by scene identity.
Our findings expose a grounding gap concealed by nominal success and provide a diagnostic framework. They further establish a concrete criterion for progress: policies should reliably follow valid changes in user intent, even when they conflict with scene-favored behavior.

\end{abstract}
\section{Introduction}

Language-conditioned robot policies promise a natural interface for specifying what a robot should do in an environment. Recent VLAs~\citep{intelligence2025pi_,bjorck2025gr00t} and WAMs~\citep{yuan2026fast,li2026causal} demonstrate strong performance across manipulation tasks, embodiments, and visual environments, commonly evaluated by task success under a given instruction and environment. However, task success alone does not establish that the instruction determined the behavior. In many robot datasets, a scene is repeatedly paired with one target object or task, so the scene alone may suffice to infer the demonstrated behavior without resolving the requested behavior. A policy can consequently achieve high success by exploiting scene--task correlations while its ability to follow language remains unclear.

Instruction following requires evidence that changing a valid instruction in a fixed scene redirects behavior toward the newly specified target. Evaluations using linguistic perturbations and modality ablations reveal brittleness and task-dependent language sensitivity in VLA policies~\citep{fei2025libero, zhou2025libero, guo2026robustness, dong2026does, shi2026vla}. More directly, LIBERO-CF and LangGap expose instruction-following failures under alternative feasible instructions in fixed scenes~\citep{fang2026vision, hou2026langgap}, while InstructMove constructs scenes where language must distinguish among multiple feasible actions~\citep{zhao2026instructmove}. Mechanistic analyses further find that prompt information can remain decodable despite limited behavioral use, with action generation relying strongly on visual inputs~\citep{grant2026not}. Building on these findings, we use \emph{scene-preserving instruction interventions} to investigate how target-selection failures relate to the encoding, visual grounding, and contribution of instruction information to action generation, linking behavioral failures to action-level responses and internal representations across VLA and world-action models.

\begin{figure}[t]
    \centering
    \vspace{-0.7em}
    \includegraphics[width=\linewidth]{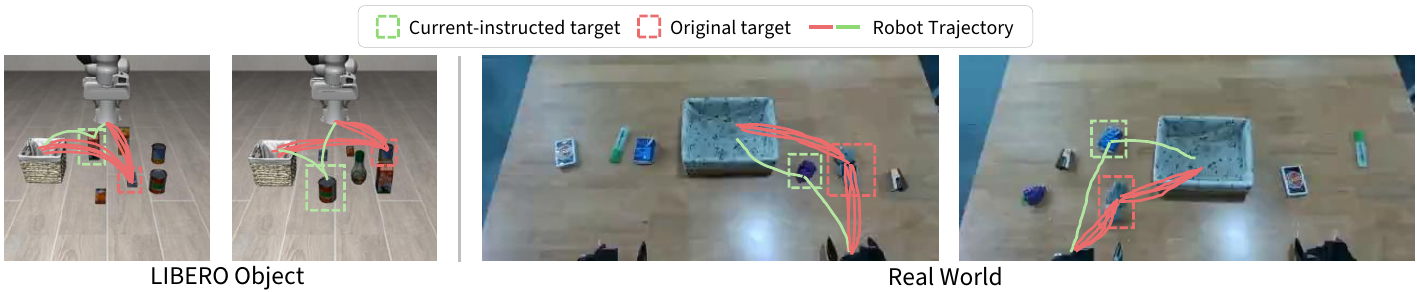}
    \vspace{-6mm}
    \caption{\textbf{Target replacement reveals a gap in instruction following.} Holding the scene fixed, we change the instruction to request another visible and reachable object. In both LIBERO-Object and real-world manipulation, policies rarely grasp the currently instructed target (\textcolor{risegreen}{dashed green trajectories}) and predominantly grasp the target named by the original instruction (\textcolor{dropred}{solid red trajectories}).}
    \label{fig:q1_qualitative}
    \vspace{-4mm}
\end{figure}

\textbf{Part I: Does nominal task success imply instruction following?}
We revisit this question across VLA and WAM policies, with additional real-world experiments. When a valid instruction requests another object in the same scene, the policies predominantly approach and manipulate the original target. We refer to this directional preference as a \emph{scene-conditioned behavioral prior}. It reveals a failure that standard success rates cannot expose: a policy may successfully reproduce the expected behavior for a familiar scene without allowing language to determine which behavior it executes.

\textbf{Part II: Is target-conflict failure caused by language insensitivity?}
This failure cannot be attributed simply to absent linguistic influence. Controlled instruction perturbations produce systematic, task-dependent changes in policy performance. A layerwise action lens shows that intermediate action predictions respond differently to target substitutions and random sentences, evidencing language sensitivity in action generation. Linear probes also recover target and scene identities from fine-tuned representations with high accuracy, so target information remains available despite incorrect object selection. Policies thus respond to language at behavioral and representational levels, yet the failure persists even when the modified instruction affects action predictions and its target remains encoded---motivating a closer look at how this information contributes to action selection.

\textbf{Part III: If target information is encoded, why does it fail to control action?}
Attention contribution analysis shows instruction tokens contribute little to action generation relative to visual inputs, and target-token attention can stay fixed on the original object even after the instruction names a different target. UMAP reveals a shift from object-based to scene-based clustering after fine-tuning; shared non-negative matrix factorization likewise identifies transitions from object-dominated to scene-dominated components in both $\pi_{0.5}$ and GR00T N1.7. High target decodability thus coexists with limited instruction contribution, misaligned visual attention, and increased scene dependence, characterizing why encoded language does not reliably guide target selection.

We describe this dissociation as a \emph{representation-to-control grounding gap}: linguistic information can be represented within a policy without reliably governing action selection. Our analyses distinguish sensitivity to language, accessibility of target information, and behavioral adherence to the instruction. These target-selection failures also extend to the evaluated world-action models, showing that visual prediction objectives alone do not ensure instruction following. Our findings motivate a stricter grounding criterion: a valid change in instruction should reliably select a different behavior within the same scene. We hope this work encourages evaluations that test such control directly and provides a diagnostic basis for investigating where encoded instructions fail to guide action. Ultimately, progress should be measured by how reliably robots act on the user's current intent, even when it conflicts with familiar scene-task associations.

\section{Related Work}

\noindent\textbf{Language-conditioned robot policies.}
Vision-language-action (VLA) models map visual observations and language instructions to robot actions. 
Recent models such as $\pi_{0.5}$ couple a pretrained
vision-language backbone with a dedicated action expert that generates
continuous action chunks through flow matching
~\citep{intelligence2025pi_}. GR00T N1 adopts a dual-system
architecture in which a vision-language module interprets visual
observations and language instructions, followed by a diffusion
transformer that generates continuous motor actions
~\citep{bjorck2025gr00t}.
World action models (WAMs) instead incorporate explicit visual-dynamics
modeling into policy learning. LingBot-VA adopts an autoregressive
diffusion framework that jointly learns future-frame prediction and
policy execution, using a shared latent space and a
Mixture-of-Transformers architecture to integrate visual and action
tokens~\citep{li2026causal}. Fast-WAM retains video co-training
during training but removes explicit future prediction at test time,
directly producing actions from the learned world representations
~\citep{yuan2026fast}. These complementary architectures motivate
our evaluation of $\pi_{0.5}$, GR00T N1.7, LingBot-VA, and FastWAM,
allowing us to examine whether instruction-grounding failures persist
across both VLA and world-action modeling paradigms.

\noindent\textbf{Evaluating and improving instruction following.}
Recent studies investigate if task success reliably reflects language
grounding. LIBERO-PRO reports persistent behaviors under corrupted
instructions~\citep{zhou2025libero}, while LIBERO-Plus evaluates robustness
to linguistic, visual, and physical perturbations~\citep{fei2025libero}.
Multilingual evaluation further reveals step-dependent language
sensitivity~\citep{dong2026does}. To test semantic control more directly,
LIBERO-CF and LangGap vary instructions within existing scene
layouts~\citep{fang2026vision,hou2026langgap}, while InstructMove introduces
semantic distractors that require language-dependent action
selection~\citep{zhao2026instructmove}. Complementary methods strengthen
language conditioning through counterfactual language--action augmentation
in navigation~\citep{glossop2026cast}, Bayesian training
objectives~\citep{lian2026langforce}, counterfactual action
guidance~\citep{fang2026vision}, and instruction augmentation coupled with
guidance against a vision-only prior~\citep{zhan2026stable}.
Building on these studies, we connect behavior under scene-preserving
instruction interventions to internal policy representations. Target substitutions test semantic redirection, whereas arbitrary nouns
and random sentences diagnose language dependence without necessarily
specifying executable alternatives.

\noindent\textbf{Language representations and behavioral influence.}
Probing studies distinguish decodable information from its use by a
model, emphasizing the need for controls when interpreting probe
accuracy~\citep{belinkov2022probing,hewitt2019designing}.
In robotics, \citet{lu2025probing} decode symbolic object and action
states from OpenVLA for monitoring within the cognitive architecture.
Using counterfactual prompts, activation injection, sparse autoencoders,
and probes, \citet{grant2026not} identify distinct VLM and action-expert
roles and report decodable prompt categories despite weak behavioral
sensitivity in $\pi_{0.5}$.
VLA-Trace combines representation analysis, attention knockouts, and
behavioral interventions to identify model-dependent control pathways
and limitations in fine-grained semantic following~\citep{shi2026vla}.
We build on these findings by pairing instruction interventions with
layer-wise analyses of target and scene information across VLA and WAM
policies, examining how instructed-target representations relate to
action selection when language and scene-associated behavior conflict.

\section{Experimental Settings}
\label{sec:experimental-settings}

\noindent\textbf{Environments and evaluation scope.}
Our experiments cover LIBERO (Spatial, Object, Goal, Long)~\citep{liu2023libero}, the clean and randomized settings of RoboTwin 2.0~\citep{chen2025robotwin}, and six real-world manipulation scenes. These environments support a broad evaluation of instruction sensitivity. For controlled target-selection experiments, we use LIBERO-Object and real-world scenes containing multiple visible and reachable objects. These settings allow the instructed target to change while preserving the initial scene, manipulation primitive, and destination, enabling us to isolate target identity from other task requirements.

\noindent\textbf{Instruction interventions.}
We consider three instruction conditions: \textsc{Original},
\textsc{Noun Replacement}, and \textsc{Random Sentence}.
\textsc{Original} retains the nominal task instruction.
\textsc{Noun Replacement} preserves the remaining instruction while modifying
object expressions, and is instantiated in two ways:
\emph{In-scene Target Replacement}, which replaces only the manipulated target
with another object present in the same scene, and
\emph{Random Noun}, which replaces all eligible object expressions in the
instruction with randomly sampled nouns.
\textsc{Random Sentence} replaces the complete instruction with a sentence
generated independently.
Details are provided in
App.~\ref{app:instruction-interventions}.

\noindent\textbf{Scene--target evaluation sets.}
For the representation analyses, we construct matched scene--target sets in
LIBERO-Object and the real world. In LIBERO-Object, each of the ten evaluation scenes contains six objects, the original target and five alternative objects present in the same scene, yielding \(10\times6=60\)
scene--target instances. Within each scene, the initial scene configuration
and robot state are fixed, and only the instructed target changes.
We construct an analogous real-world set with six scenes and six objects per scene, yielding \(6\times6=36\) instances.
The complete scene--target sets are provided in
App.~\ref{app:scene-target-sets}.

\noindent\textbf{Models.}
We evaluate two vision-language-action policies, $\pi_{0.5}$~\citep{intelligence2025pi_} and GR00T N1.7~\citep{bjorck2025gr00t}, and two world-action models, FastWAM~\citep{yuan2026fast} and LingBot-VA~\citep{li2026causal}. All four models are evaluated in simulation, while real-world experiments use $\pi_{0.5}$ and GR00T N1.7. Behavioral evaluations use closed-loop rollouts with frozen policy parameters. Model-specific observation interfaces, action spaces, episode horizons, action chunk length, instruction instances, and rollout counts are provided in the appendix.

\begin{table*}[t]
    \centering
    \small
    \caption{\textbf{Task completion under controlled target replacement.}
The \emph{current instructed target} is another object in the same scene requested by the modified instruction; the \emph{original target} is named by the original instruction. ``Tend'' denotes a committed approach or manipulation attempt, and ``Success'' denotes successful task completion. All values are percentages (\%). Policies predominantly retain the original target despite the modified instruction.}
\vspace{-2mm}
    \label{tab:target_selection}

    \setlength{\tabcolsep}{5pt}
    \renewcommand{\arraystretch}{1.15}

    \begin{tabular*}{\linewidth}{
        @{\extracolsep{\fill}} llcccc @{}
    }
        \toprule
        \multirow{2}{*}{\textbf{Environment}}
        & \multirow{2}{*}{\textbf{Model}}
        & \multicolumn{2}{c}{\textbf{Current instructed target}}
        & \multicolumn{2}{c}{\textbf{Original target}} \\
        \cmidrule(lr){3-4}
        \cmidrule(lr){5-6}
        & & Tend & Success & Tend & Success \\
        \midrule

        \multirow{4}{*}{\textit{LIBERO-Object}}
        & $\pi_{0.5}$ & $13.0$ & $11.5$ & $86.5$ & $69.5$ \\
        & GR00T N1.7  & $1.3$  & $0.5$  & $96.5$ & $90.9$ \\
        & FastWAM     & $0.5$  & $0.0$  & $98.3$ & $95.0$ \\
        & LingBot-VA  & $1.8$  & $1.3$  & $96.3$ & $84.4$ \\

        \addlinespace[2pt]
        \midrule
        \addlinespace[2pt]

        \multirow{2}{*}{\textit{Real World}}
        & $\pi_{0.5}$ & $10.8$ & $5.8$ & $80.8$ & $74.2$ \\
        & GR00T N1.7  & $4.2$  & $3.3$ & $94.2$ & $90.8$ \\

        \bottomrule
    \end{tabular*}
    \vspace{-4mm}
\end{table*}
\section{Does Nominal Task Success Imply Instruction Following?}
\label{sec:q1}

\begin{questionbox}
\textbf{Guiding question Q1.}
\textit{When the instruction requests another valid target in the same scene, does the policy follow the current instruction or retain the original target?}
\end{questionbox}

Successful execution cannot distinguish instruction following from task inference when the scene and instruction suggest the same behavior. Following prior evaluations of alternative feasible instructions in fixed scenes~\citep{fang2026vision,hou2026langgap}, we revisit this ambiguity through controlled target replacement. We change the instructed in the text prompt while preserving the initial scene and other robot configurations. 

\subsection{Target Replacement in LIBERO-Object and the Real World}
\label{subsec:q1_results}

We evaluate target replacement in LIBERO-Object using $\pi_{0.5}$, GR00T N1.7, FastWAM, and LingBot-VA, and in six real-world scenes using the two VLA policies. Alternative targets are visible and reachable; in LIBERO-Object, each manipulable object is also represented in the training set. We report committed approaches or manipulation attempts (\emph{Tend}) and
successful task execution (\emph{Success}). \emph{Tend} records whether the
policy makes a committed approach or manipulation attempt toward a target,
whereas \emph{Success} requires successfully finish the task.

Across the four models in LIBERO-Object, $86.5\%$--$98.3\%$ of trials are directed toward the original target, compared with $0.5\%$--$13.0\%$ toward the current instructed target (Tab.~\ref{tab:target_selection}). Original-target task completion reaches $69.5\%$--$95.3\%$, while current-target completion remains at $0.0\%$--$11.5\%$. The same preference persists in physical execution: original-target pickup rates are $74.2\%$ and $90.8\%$ for $\pi_{0.5}$ and GR00T N1.7, respectively, compared with $5.8\%$ and $3.3\%$ for the current target.

These results corroborate previously reported counterfactual failures~\citep{fang2026vision,hou2026langgap} and extend the evidence to the evaluated world-action models. The policies frequently manipulate the object that is no longer requested, as illustrated in Fig.~\ref{fig:q1_qualitative}, motivating closer analysis of how the modified instruction is processed.

\begin{takeawaybox}
\textbf{Answer to Q1.}
\textit{\textbf{Nominal task success does not establish instruction following.} Across the evaluated policies, valid target replacement predominantly leaves behavior directed toward the original target, despite successful manipulation.}
\end{takeawaybox}


\begin{table*}[t]
    \centering
    \small
    \renewcommand{\arraystretch}{1.15}
    \setlength{\tabcolsep}{4pt}
    \caption{\textbf{Task performance under instruction corruption.}
Main values denote success rates (\%); colored subscripts indicate relative decreases ({\color{dropred}$\downarrow$}) or increases ({\color{risegreen}$\uparrow$}) from the original-instruction baseline (\%). The results demonstrate that instruction corruption affects performance, with sensitivity varying across tasks and evaluation conditions.}

    \label{tab:instruction_corruption}
    \resizebox{\linewidth}{!}{%
    \begin{tabular}{@{}lllcccc@{}}
        \toprule
        \multirow{2}{*}{\textbf{Benchmark}}
        & \multirow{2}{*}{\textbf{Task / Setting}}
        & \multirow{2}{*}{\textbf{Instruction}}
        & \multicolumn{2}{c}{\textbf{VLA}}
        & \multicolumn{2}{c}{\textbf{WAM}} \\
        \cmidrule(lr){4-5}
        \cmidrule(lr){6-7}
        & &
        & $\pi_{0.5}$
        & GR00T N1.7
        & FastWAM
        & LingBot-VA \\
        \midrule

        \multirow{12}{*}{\textit{LIBERO}}
        & \multirow{3}{*}{Spatial}
        & Original
        & \srorig{98.8}
        & \srorig{97.7}
        & \srorig{98.2}
        & \srorig{98.5} \\
        & & Random noun
        & \srup{99.0}{0.2}
        & \srdown{96.6}{1.1}
        & \srdown{73.6}{25.1}
        & \srdown{90.4}{8.2} \\
        & & Random sentence
        & \srdown{66.4}{32.8}
        & \srdown{53.2}{45.5}
        & \srdown{37.4}{61.9}
        & \srdown{75.2}{23.7} \\
        \cmidrule(lr){2-7}

        & \multirow{3}{*}{Object}
        & Original
        & \srorig{98.2}
        & \srorig{98.5}
        & \srorig{100.0}
        & \srorig{99.6} \\
        & & Random noun
        & \srdown{96.6}{1.6}
        & \srdown{94.4}{4.1}
        & \srdown{92.6}{7.4}
        & \srdown{83.4}{16.3} \\
        & & Random sentence
        & \srdown{64.9}{33.9}
        & \srdown{52.8}{46.4}
        & \srdown{87.0}{13.0}
        & \srdown{49.2}{50.6} \\
        \cmidrule(lr){2-7}

        & \multirow{3}{*}{Goal}
        & Original
        & \srorig{98.0}
        & \srorig{97.5}
        & \srorig{97.0}
        & \srorig{97.2} \\
        & & Random noun
        & \srdown{81.5}{16.8}
        & \srdown{75.2}{22.9}
        & \srdown{49.0}{49.5}
        & \srdown{61.4}{36.8} \\
        & & Random sentence
        & \srdown{11.7}{88.1}
        & \srdown{4.0}{95.9}
        & \srdown{4.8}{95.1}
        & \srdown{12.0}{87.7} \\
        \cmidrule(lr){2-7}

        & \multirow{3}{*}{Long}
        & Original
        & \srorig{92.4}
        & \srorig{94.4}
        & \srorig{95.2}
        & \srorig{98.5} \\
        & & Random noun
        & \srup{94.2}{2.0}
        & \srdown{87.4}{7.4}
        & \srdown{82.2}{13.7}
        & \srdown{89.6}{9.0} \\
        & & Random sentence
        & \srdown{78.6}{14.9}
        & \srdown{59.6}{36.8}
        & \srdown{47.6}{50.0}
        & \srdown{79.8}{19.0} \\
        \midrule

        \multirow{6}{*}{\textit{RoboTwin 2.0}}
        & \multirow{3}{*}{Clean}
        & Original
        & \srorig{72.8}
        & \srorig{69.5}
        & \srorig{91.9}
        & \srorig{92.9} \\
        & & Random noun
        & \srdown{60.3}{17.2}
        & \srdown{58.2}{16.2}
        & \srdown{81.6}{11.2}
        & \srdown{82.6}{11.1} \\
        & & Random sentence
        & \srdown{48.9}{32.9}
        & \srdown{48.4}{30.4}
        & \srdown{51.0}{44.5}
        & \srdown{59.6}{35.8} \\
        \cmidrule(lr){2-7}

        & \multirow{3}{*}{Randomized}
        & Original
        & \srorig{71.3}
        & \srorig{67.4}
        & \srorig{91.8}
        & \srorig{91.6} \\
        & & Random noun
        & \srdown{56.1}{21.3}
        & \srdown{48.9}{27.4}
        & \srdown{76.6}{16.5}
        & \srdown{76.4}{16.6} \\
        & & Random sentence
        & \srdown{47.0}{34.2}
        & \srdown{42.1}{37.5}
        & \srdown{38.8}{57.7}
        & \srdown{48.6}{46.9} \\
        \bottomrule
    \end{tabular}%
    }
    \vspace{-5mm}
\end{table*}

\section{Is This Failure Caused by Language Insensitivity?}
\label{sec:q2}

\begin{questionbox}
\textbf{Guiding question Q2.}
\textit{Do policies remain sensitive to language and encode the currently instructed target despite failing to follow it?}
\end{questionbox}

The original-target preference in Sec.~\ref{sec:q1} does not establish whether the policies ignore the modified instruction. We examine language sensitivity at three levels: changes in task success under instruction perturbations, changes in intermediate action predictions measured by a layerwise action lens, and target decodability measured by linear probes. These analyses test whether instruction-following failures persist even when language affects action prediction and the currently instructed target remains recoverable from the policy's representations.

\subsection{Instruction Perturbations Reveal Behavioral Sensitivity}
\label{sec:q2.1}
We evaluate the instruction-corruption conditions defined in Sec.~\ref{sec:experimental-settings}:
\textsc{Original}, \textsc{Any Noun}, and \textsc{Random Sentence}.
All four policies are evaluated on LIBERO and RoboTwin 2.0 under clean and
randomized settings, with success measured against the original task.

\textbf{LIBERO:} Random sentences reduce success across all four models and evaluated suites (Tab.~\ref{tab:instruction_corruption}). Noun replacement generally produces smaller, task-dependent reductions. For $\pi_{0.5}$ and GR00T N1.7, its effect is limited on Spatial, Object, and Long, but larger on Goal. Both world-action models likewise exhibit their largest noun-replacement reductions on LIBERO-Goal.
\textbf{RoboTwin:} Both interventions reduce success for every model in the clean and randomized settings. Random sentences consistently have a larger effect than noun replacement.
These results show that language affects task performance, with sensitivity varying across tasks and evaluation conditions.

\subsection{Instruction Changes Affect Intermediate Action Predictions}
\label{sec:action_lens}
Inspired by the logit lens~\citep{nostalgebraist2020}, we examine action-generation sensitivity using a layerwise action lens. At each layer $l$ and denoising step $s$, we apply the policy's action readout to intermediate hidden states in normalized action coordinates. These readouts diagnose intermediate representations rather than provide calibrated predictions of executed actions. For the same visual observation, we compare each intervention $q$ with the original instruction at the same layer and step:
\begin{equation}
D_q^{(s,l)}
=
\frac{1}{d}
\left\|
\hat{\mathbf{a}}_q^{(s,l)}
-
\hat{\mathbf{a}}_{\mathrm{orig}}^{(s,l)}
\right\|_2^2,
\end{equation}
where $d$ is the number of actions within an episode. We compare noun replacement with an out-of-scene noun or another visible object, and complete instruction replacement with a random sentence. Details are provided in App.~\ref{app:logit_lens}.

\begin{figure*}[t]
    \centering

    \includegraphics[width=\textwidth]{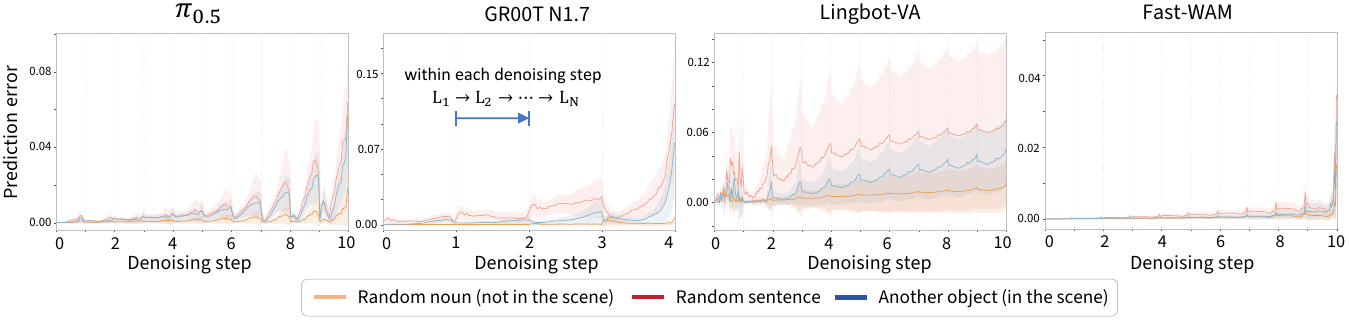}

    \caption{\textbf{Layerwise action predictions under instruction interventions.}
Curves show mean squared deviations from original-instruction predictions at matched layers and denoising steps. Random sentences (red) generally produce the largest deviations, followed by in-scene target replacements (blue) and out-of-scene nouns (orange), indicating language sensitivity within generation.}

    \label{fig:action_lens}
    \vspace{-5mm}
\end{figure*}

Across all four models, Fig.~\ref{fig:action_lens} shows a consistent overall ordering: out-of-scene nouns yield the smallest deviations, followed by in-scene replacements and random sentences. Separation generally increases toward later denoising steps, with variation across layers and models. These results show that instruction changes affect intermediate action predictions, complementing their effects on task performance in Sec.~\ref{sec:q2.1}. However, deviation does not measure movement toward the instructed target: rollouts in Sec.~\ref{sec:q1} predominantly retain the original target. We next test whether the current target's identity remains recoverable from internal representations.

\subsection{The Current Target Remains Linearly Decodable}
\label{sec:linear_probe}

We probe whether the currently instructed target remains accessible in the
internal representations of $\pi_{0.5}$ and GR00T N1.7 before and after
fine-tuning. Using the matched scene--target sets from LIBERO-Object and the
real world, we train separate linear probes to predict target identity and
scene identity from instruction representations. Full feature-extraction,
label-construction, training, and evaluation details are provided in
App.~\ref{app:linear_probe}.
\begin{figure*}[t]
    \centering
    \includegraphics[width=.86\linewidth]{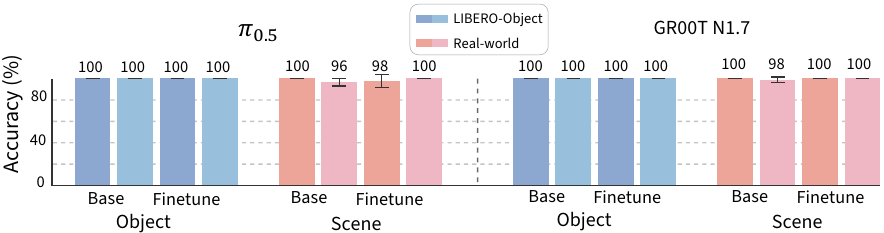}
    \vspace{-3mm}
    \caption{\textbf{Linear probing of target and scene identity.}
Probes on mean-pooled instruction-token representations achieve near-ceiling accuracy in $\pi_{0.5}$ and GR00T before and after fine-tuning on LIBERO-Object and real-world data, showing that both identities remain accessible despite instruction-following failures.}
    \label{fig:linear_probe}
\vspace{-4mm}
\end{figure*}
Fig.~\ref{fig:linear_probe} shows near-ceiling decodability in both models
and environments. On a test set, target identity is recovered with essentially perfect
accuracy before and after fine-tuning, while scene identity is also decoded
with near-ceiling accuracy.

The currently instructed target therefore remains linearly accessible
despite the unreliable target selection observed in
Sec.~\ref{sec:q1}. This rules out complete loss of target identity from the
probed representations as an explanation for the behavioral failure.

\begin{takeawaybox}
\textbf{Answer to Q2.}
\textit{\textbf{The failure cannot be explained simply by language insensitivity or the absence of target information.} Instruction perturbations affect performance, and the current target remains highly decodable in the probed policies, despite their unreliable target following.}
\end{takeawaybox}


\section{Why Does Encoded Language Fail to Control Action?}
\label{sec:q3}

\begin{questionbox}
\textbf{Guiding question Q3.}
\textit{If the current target remains encoded, how can visual attention and action generation remain misaligned with the instruction, and what changes in representational organization accompany this failure?}
\end{questionbox}

Sec.~\ref{sec:q2} shows that instruction changes affect task performance and intermediate action predictions, while the current target remains linearly decodable in the probed policies. We now examine how this information participates in action generation and relates to visual grounding. We first quantify the relative contribution of instruction tokens to action generation.
Then UMAP and shared NMF characterize how target- and scene-dependent organization changes after fine-tuning. These analyses provide complementary diagnostics of the gap between target encoding and behavioral control.

\subsection{Instruction Tokens Contribute Weakly to Action Generation}
\label{sec:action_contribution}

Target decodability does not establish how strongly instruction tokens contribute to action generation. We quantify their relative contribution to the action expert by aggregating attention-weighted value vectors after output projection~\citep{kobayashi2020attention,song2026elastic}. For action-query token $i$ and source token $j$, we compute
\begin{equation}
e_{ij}
=
\sum_{\ell=1}^{L}
\sum_{h=1}^{H_\ell}
A_{hij}^{\ell}
\left(v_{hj}^{\ell}W_h^{O,\ell}\right),
\qquad
s_{ij} = \frac{\left\|e_{ij}\right\|_2} {\sum_{j'}\left\|e_{ij'}\right\|_2}. \label{eq:action_contribution}
\end{equation}
where $A_{hij}^{\ell}$ is the attention probability, $v_{hj}^{\ell}$ the value vector, and $W_h^{O,\ell}$ the head's output projection. Normalization follows aggregation across all layers and heads. We then sum $s_{ij}$ by source modality and aggregate over active action-query tokens.

\begin{figure*}[t]
    \centering
    \includegraphics[width=\linewidth]{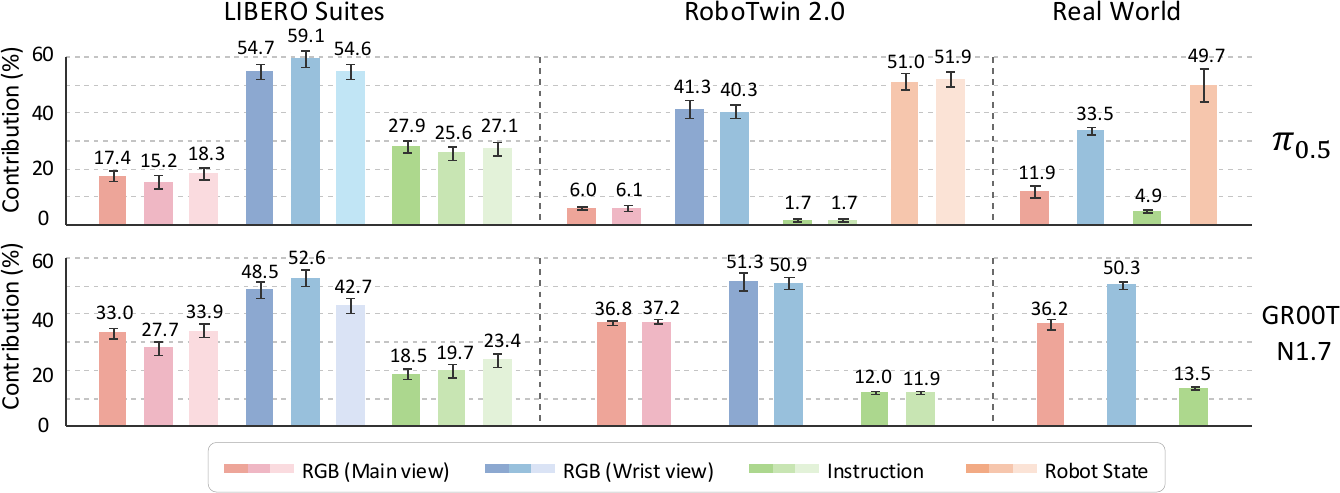}
\caption{\textbf{Modality contributions to action generation.}
Instruction tokens contribute relatively weakly to action generation in both $\pi_{0.5}$ and GR00T N1.7, with visual or robot-state inputs accounting for larger shares across simulation and real-world settings.}
    \label{fig:action_contribution}
    \vspace{-6mm}
\end{figure*}

Instruction tokens account for a smaller share of the measured contribution than the dominant visual or state inputs (Fig.~\ref{fig:action_contribution}). For $\pi_{0.5}$, this share is $25.6\%$--$27.9\%$ on LIBERO, approximately $1.7\%$ on RoboTwin~2.0, and $9.7\%$ in the real world. The corresponding values for GR00T N1.7 are $18.5\%$--$23.4\%$, approximately $12\%$, and $13.5\%$.
Thus, high target decodability coexists with a relatively small instruction contribution under this attention-based measure. 
\subsection{Target Replacement Does Not Reliably Redirect Visual Attention}
\label{sec:target_attention}

\begin{wrapfigure}{r}{0.50\textwidth}
    \centering
    \vspace{-3mm}
    \includegraphics[width=\linewidth]{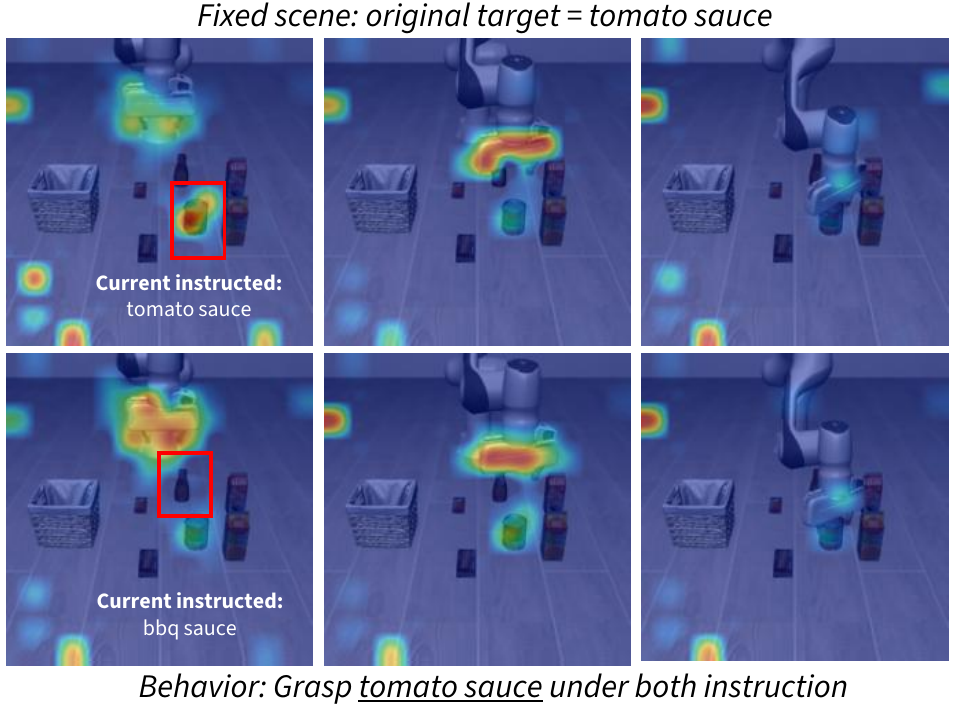}
    \caption{\textbf{Target-token visual attention under target replacement.}
When the instructed target changes from tomato sauce to BBQ sauce, attention remains focused on the original tomato sauce, which the policy grasps under both instructions. Red boxes mark the currently instructed target.}

    \vspace{-7mm}
    \label{fig:attn_map}
\end{wrapfigure}

We examine whether attention from instructed-target tokens aligns with the corresponding object using the LIBERO-Object evaluation set 
Fig.~\ref{fig:attn_map} shows a representative $\pi_{0.5}$ example in which the instructed target changes from \emph{tomato sauce} to \emph{BBQ sauce}. 

This observation reveals a mismatch between target encoding and visual grounding: the current target remains decodable, yet target-token attention can favor the original object. This pattern accompanies the behavioral failures in Sec.~\ref{sec:q1}, without establishing a causal role for attention. We next examine whether fine-tuning also changes how instruction representations are organized around target and scene identity.

\subsection{Fine-Tuning Shifts Instruction Representations toward Scene Identity}
\label{sec:umap}
We examine how fine-tuning changes the organization of instruction
representations in $\pi_{0.5}$. Using the matched LIBERO-Object
scene--target evaluation set, we visualize layer-wise instruction
representations from the base and fine-tuned checkpoints with UMAP.
Details are provided in App.~\ref{app:umap_details}.

\begin{figure*}[t]
    \centering
    \includegraphics[width=\textwidth]{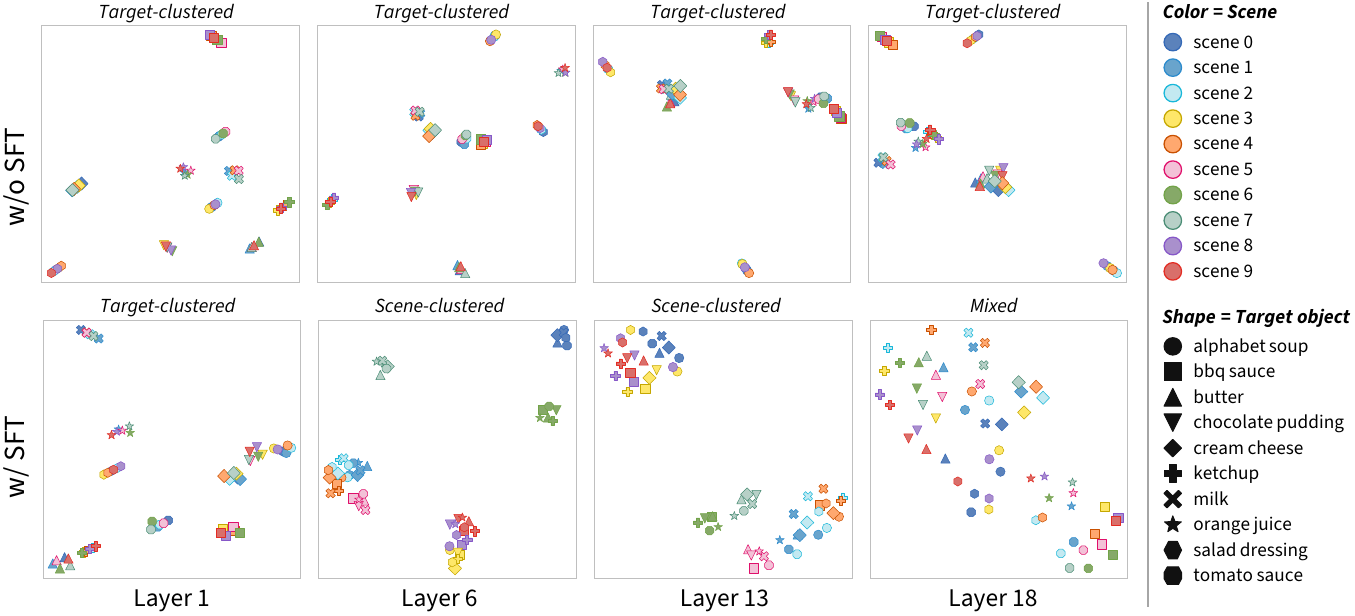}
    \vspace{-2mm}
    \caption{\textbf{UMAP projections of $\pi_{0.5}$ instruction representations.}
Columns show layers 1, 6, 13, and 18; rows compare Base and SFT.
Shape denotes target identity; color denotes scene.
Base representations cluster by target. After fine-tuning, the first layer retains target clustering, intermediate layers group by scene, and the final layer is more dispersed with residual target grouping.}

    \label{fig:umap_all}
    \vspace{-4mm}
\end{figure*}

As shown in Fig.~\ref{fig:umap_all}, before fine-tuning, the projections
show clear target-based clustering across the inspected layers.
Fine-tuning substantially reorganizes this geometry. The early layer
retains clear target-based clustering, intermediate layers become
predominantly organized by scene identity, and the final layers are
more dispersed while retaining a tendency toward target-based grouping.
Together with the probing results, these visualizations suggest that fine-tuning changes how target information is organized, without erasing it.

\subsection{Shared Semi-NMF Reveals a Shift toward Scene-Dependent Organization}
\label{sec:shared_nmf}

We quantify the reorganization observed in UMAP using shared semi-NMF~\citep{ding2008convex}. We jointly factorize matched base and fine-tuned instruction representations:
\begin{equation}
\begin{bmatrix}
X_{\mathrm{base}}\\
X_{\mathrm{FT}}
\end{bmatrix}
\approx
\begin{bmatrix}
A_{\mathrm{base}}\\
A_{\mathrm{FT}}
\end{bmatrix}B^\top,
\qquad
A_{\mathrm{base}},A_{\mathrm{FT}}\geq0,
\label{eq:shared_semi_nmf}
\end{equation}
\begin{wrapfigure}{r}{0.50\textwidth}
    \centering
    \vspace{-2mm}
    \includegraphics[width=\linewidth]{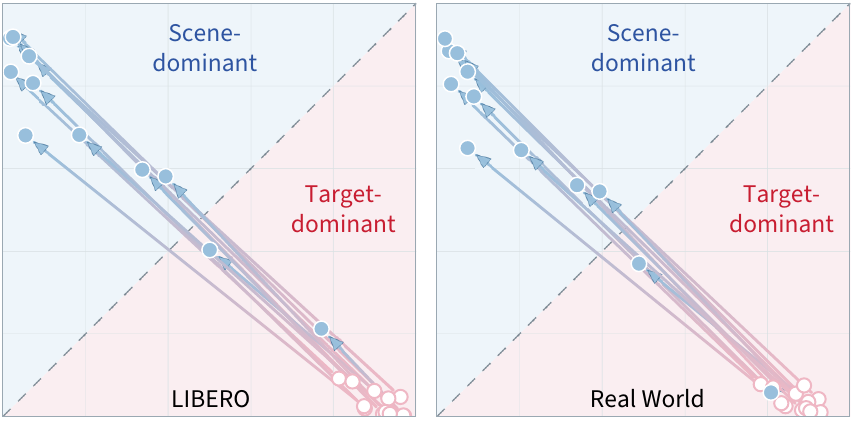}
    \vspace{-5mm}
    \caption{\textbf{Shared semi-NMF components before and after fine-tuning.}
Arrows connect components before (open circles) and after fine-tuning (filled circles). Across both settings, components shift from target-dominated to scene-dominated organization, with greater scene-unique and lower target-unique variance.}
    \label{fig:semi_nmf}
    \vspace{-5mm}
\end{wrapfigure}
where the shared signed basis $B$ aligns components across checkpoints, and $A$ contains their condition-level activations.

For each component, we fit scene-only, target-only, and additive scene--target models. Their explained variances, $R_S^2$, $R_T^2$, and $R_{ST}^2$, yield the unique contributions of scene and target identity:
\begin{equation}
U_{\mathrm{scene}}=R_{ST}^2-R_T^2,
\qquad
U_{\mathrm{target}}=R_{ST}^2-R_S^2.
\label{eq:semi_nmf_unique_variance}
\end{equation}

Each term measures the additional variance explained by one factor after controlling for the other. Implementation and variance-partitioning details are provided in App.~\ref{app:semi_nmf_details}.

In the $(U_{\mathrm{target}},U_{\mathrm{scene}})$ plane, all 12 components shift upper-left at every analyzed layer from layer~2 onward, indicating increased scene-unique and decreased target-unique variance after fine-tuning. 

Representative results in Fig.~\ref{fig:semi_nmf} illustrate the shift from target-dominated to scene-dominated organization in LIBERO and the real-world setting. Preserved target information therefore coexists with scene-dominated organization.

Although both scene and target identities remain linearly decodable (as described in Sec.~\ref{sec:linear_probe}), fine-tuning shifts the shared components from target-dominated toward scene-dominated organization. Target information is preserved, but becomes less prominent in the variation captured by these components. This helps reconcile high probe accuracy with instruction-following failures and is consistent with the preference for the original target, without establishing a causal link.

\begin{takeawaybox}
\textbf{Answer to Q3.}
\textit{\textbf{Target information remains encoded, but its participation in action generation and alignment with the instructed object are limited.} Instruction tokens account for a relatively small share of the measured attention contribution to action generation, target-token visual attention can remain focused on the original object, and fine-tuning shifts representational organization toward scene identity. These findings characterize how preserved target information coexists with unreliable instruction following, without establishing a causal mechanism for the failure.}
\end{takeawaybox}

\section{Conclusion}
\label{sec:conclusion}

We investigate instruction grounding through a progression from behavioral failures to language sensitivity, target encoding, and participation in action generation. Across the evaluated policies, changed instructions affect computation and preserve recoverable target information, yet often fail to redirect target selection. The internal analyses reveal weak instruction contributions, persistent attention to the original object, and a shift toward scene-dependent representational organization. This evidence places the use of encoded language at the center of the grounding problem. It motivates evaluation and training that explicitly require instructions to select among competing valid behaviors within the same scene.

\bibliography{ref}
\bibliographystyle{iclr2027_conference}

\appendix

\section*{Appendix}

\begin{enumerate}
    \item \hyperref[app:general_settings]{General Experimental Settings}
    -- Model sources and implementations, benchmark descriptions, and evaluation protocols.

    \item \hyperref[app:instruction-interventions]{Instruction Intervention Details}
    -- Construction of \textsc{Noun Replacement}
    (\emph{In-scene Target Replacement} and \emph{Random Noun})
    and \textsc{Random Sentence}.

    \item \hyperref[app:scene-target-sets]{Details on Scene--Target Evaluation Sets}
    -- Construction of the matched LIBERO-Object and real-world scene--target sets, including scenes, instructed-target sets, and instruction templates.

    \item \hyperref[app:linear_probe]{Linear Probe Settings}
    -- Scene- and target-classification tasks, feature extraction, one-hot labels, and ridge-regression protocol.

    \item \hyperref[app:logit_lens]{Layer-wise Action Lens Analysis}
    -- Paired counterfactual replay, model-specific intermediate action readouts, and MSE-based comparison.

    \item \hyperref[app:semi_nmf_details]{NMF Analysis}
    -- Factorization setup, representation preprocessing, shared components, and quantitative analysis.

    \item \hyperref[app:detailed_literature_review]{Literature Review}
      -- Detailed literature review on Vision--action models, vision--language--action models, world--action models, and agentic robotics.
\end{enumerate}
\section{General Experimental Settings}
\label{app:general_settings}

\subsection{Model Sources and Implementations}

We evaluate four language-conditioned robot policies 
two model
families: two vision-language-action (VLA) policies, $\pi_{0.5}$~\citep{intelligence2025pi_}
and GR00T N1.7~\citep{bjorck2025gr00t}, and two world-action models (WAMs),
FastWAM~\citep{yuan2026fast} and LingBot-VA~\citep{li2026causal}.
All behavioral evaluations are conducted in closed loop with frozen policy
parameters. We retain the observation preprocessing, action representation,
and inference interface of each public implementation rather than converting
all models into a shared policy architecture.

\paragraph{Source repositories.}
The model implementations are based on the following public repositories:
\begin{itemize}
    \item $\pi_{0.5}$: Physical Intelligence OpenPI
    (\href{https://github.com/Physical-Intelligence/openpi}{GitHub}).

    \item GR00T N1.7: NVIDIA Isaac-GR00T
    (\href{https://github.com/NVIDIA/Isaac-GR00T}{GitHub}).

    \item FastWAM
    (\href{https://github.com/yuantianyuan01/FastWAM}{GitHub}).

    \item LingBot-VA
    (\href{https://github.com/Robbyant/lingbot-va}{GitHub}).

    \item XPolicyLab
    (\href{https://github.com/XPolicyLab/XPolicyLab}{GitHub}),
    used for our RoboTwin training pipeline for
    $\pi_{0.5}$ and GR00T N1.7.
\end{itemize}

\subsection{LIBERO Benchmark}

LIBERO~\citep{liu2023libero} is a language-conditioned manipulation
benchmark designed to study knowledge transfer across manipulation tasks.
We evaluate the four commonly used ten-task suites:
\textsc{LIBERO-Spatial}, \textsc{LIBERO-Object},
\textsc{LIBERO-Goal}, and \textsc{LIBERO-Long}.

The suites emphasize different forms of generalization.
\textsc{Spatial} varies spatial relations while largely preserving object
semantics; \textsc{Object} varies the manipulated objects;
\textsc{Goal} varies task goals; and \textsc{Long} contains longer-horizon,
multi-stage manipulation tasks. Each task is associated with a natural-language
instruction and a benchmark-defined success predicate.

The LIBERO simulator uses a Franka Panda robot and provides RGB observations
from an external agent-view camera and an eye-in-hand wrist camera together
with robot proprioception. The low-level control action
is seven-dimensional, consisting of end-effector translation, end-effector
rotation, and gripper control. We use the benchmark-provided initial states
and native task-success predicates.

For all instruction-intervention experiments, the initial environment state
and robot state are held fixed across instruction conditions. Thus, the
\textsc{Original}, \textsc{Noun Replacement}, and
\textsc{Random Sentence} conditions differ only in the language input.

\subsection{LIBERO Evaluation Protocols}

Because the four policies expose different action-chunking and inference
interfaces, we retain the evaluation protocol associated with each public
implementation.
For each model, we use a checkpoint fine-tuned on the full LIBERO benchmark.
When such a checkpoint is not publicly available, we fine-tune the model
ourselves following the official training guidelines and recommended settings.
Importantly, the evaluation protocol for a given model is kept identical
across all instruction conditions.

\subsection{RoboTwin 2.0 Benchmark}

RoboTwin 2.0~\citep{chen2025robotwin} is a simulation benchmark for
generalizable bimanual robotic manipulation. We evaluate all four policies
on the 50-task benchmark using the ALOHA-AgileX dual-arm embodiment.

We consider two evaluation conditions. The \textsc{Clean} condition uses
the nominal task configuration, whereas the \textsc{Randomized} condition
introduces benchmark-defined domain randomization.
RoboTwin 2.0 randomizes multiple visual and physical factors, including
background and tabletop textures, scene clutter, illumination, tabletop
height, and camera viewpoint. These variations test whether a policy trained
on demonstrated manipulation behaviors remains robust to changes in the
surrounding scene.

The RoboTwin environment provides observations from a global head camera
and two wrist cameras together with the bimanual robot state. Policies predict actions for the ALOHA-AgileX
embodiment through their corresponding RoboTwin policy adapters.

\subsection{RoboTwin Training and Evaluation Protocol}

For $\pi_{0.5}$ and GR00T N1.7, the RoboTwin checkpoints used in this
paper are trained by us following the default training pipelines provided
by XPolicyLab. FastWAM and LingBot-VA are evaluated
through their corresponding public RoboTwin implementations.

RoboTwin 2.0 uses task-specific evaluation horizons rather than a single
global maximum episode length. In evaluation mode,
each task reads its control-step limit from the benchmark configuration. We retain these benchmark-defined limits.

Following the benchmark evaluation protocol, each
task is evaluated with 100 rollouts. An episode terminates when either the
task-specific success condition is satisfied or its corresponding control-step
limit is reached. The benchmark success predicate is used directly without
introducing an additional task-completion heuristic.

Within each model, the same simulator seeds, task configurations, episode
limits, observation preprocessing, and action-execution parameters are used
for the \textsc{Original}, \textsc{Noun Replacement}, and
\textsc{Random Sentence} conditions. Only the language instruction is
modified.

\subsection{Matched Evaluation under Instruction Interventions}

Across both LIBERO and RoboTwin
2.0, instruction interventions are evaluated under
matched environmental conditions. Let $s_0$ denote an initial simulator
state and $\ell$ the language instruction. For a given evaluation instance,
we keep $s_0$ fixed and replace only $\ell$. Consequently, comparisons between
\textsc{Original}, \textsc{Noun Replacement}, and
\textsc{Random Sentence} isolate the effect of language from variation
in environment initialization.

Unless otherwise stated, success is measured using the native benchmark
predicate and against the task specified by the original evaluation
instance. Policies remain frozen throughout evaluation; no online adaptation
or parameter update is performed.
\section{Instruction Intervention Details}
\label{app:instruction-interventions}

This appendix provides the implementation details for the instruction
interventions defined in Sec.~\ref{sec:experimental-settings}.

\subsection{Noun Replacement}
\label{app:noun-replacement}

\paragraph{\emph{In-scene Target Replacement}.}
For LIBERO-Object, we identify the manipulated target expression in the
instruction. Each other manipulable object present in the same scene is used
in turn as a replacement for the manipulated target, while the remaining
instruction is left unchanged.

For example,
\begin{quote}
\small
\emph{pick up the \underline{ketchup} and place it in the basket}
$\rightarrow$
\emph{pick up the \underline{milk} and place it in the basket}.
\end{quote}

\paragraph{\emph{Random Noun}.}
For this variant, all eligible object expressions in the instruction are
identified and replaced with randomly sampled nouns.

For LIBERO, eligible object expressions are identified by matching complete
object names in the instruction against the 45-item vocabulary in
Table~\ref{tab:language-perturbation-pools}.
All matched expressions are replaced independently with nouns sampled uniformly 
from the 99-item replacement vocabulary.

For RoboTwin 2.0, eligible object expressions are resolved from the
task-instruction template, \texttt{episode\_info}, and the corresponding
object-description metadata. This resolution includes episode-specific and
multiword object descriptions. Every resolved eligible object expression is
replaced, with each replacement noun sampled independently and uniformly from
the same 99-item replacement vocabulary.

\begin{table*}[t]
\centering
\caption{Lexical pools used for the \emph{Random Noun} and
\textsc{Random Sentence} interventions.}
\label{tab:language-perturbation-pools}
\scriptsize
\setlength{\tabcolsep}{4pt}
\begin{tabular}{p{0.17\textwidth} p{0.78\textwidth}}
\toprule
Pool & Members \\
\midrule

LIBERO object vocabulary used for eligibility (45) &
bowl, plate, ramekin, cup, mug, box, cabinet, drawer, basket, book, butter,
cheese, chocolate, cream, sauce, can, bottle, pan, pot, lid, tray, rack,
shelf, table, desk, chair, cloth, towel, bag, container, toy, block, cube,
ball, hammer, brush, fork, spoon, knife, glass, jar, kettle, pitcher, bin,
bucket. \\

\addlinespace

Replacement and random-sentence nouns (99) &
cat, dog, bird, fish, horse, lion, tiger, bear, wolf, rabbit, deer, eagle,
shark, whale, frog, snake, turtle, rock, stone, tree, leaf, cloud, river,
mountain, beach, forest, lake, desert, island, volcano, cave, valley, lamp,
clock, mirror, pillow, blanket, carpet, curtain, vase, candle, picture,
statue, phone, remote, cable, apple, banana, orange, grape, pizza, bread,
rice, soup, cookie, cake, sandwich, salad, pasta, egg, milk, wrench, nail,
screw, drill, rope, chain, key, lock, pipe, wire, tape, glue, paint, wheel,
gear, shirt, hat, shoe, sock, coat, glove, belt, scarf, coin, stamp, ticket,
card, map, flag, sign, label, button, switch, ring, medal, trophy, balloon,
umbrella. \\

\addlinespace

Random-sentence verbs (10) &
observe, locate, inspect, examine, survey, approach, avoid, ignore, study,
scan. \\

\addlinespace

Random-sentence adjectives (10) &
curious, ancient, shiny, purple, enormous, tiny, mysterious, hollow, rusty,
invisible. \\

\bottomrule
\end{tabular}
\end{table*}

\subsection{Random Sentence Generation}
\label{app:random-sentence}

For \textsc{Random Sentence}, the original instruction is discarded. One of
the ten templates in Table~\ref{tab:random-sentence-templates} is sampled
uniformly. Each noun, verb, and adjective slot is filled independently and
uniformly, with replacement, from the corresponding lexical pool in
Table~\ref{tab:language-perturbation-pools}. No task-dependent,
scene-dependent, semantic, or grammatical filtering is applied.

\begin{table*}[t]
\centering
\caption{Templates used to construct the
\textsc{Random Sentence} intervention.}
\label{tab:random-sentence-templates}
\small
\setlength{\tabcolsep}{5pt}
\begin{tabular}{r p{0.90\textwidth}}
\toprule
No. & Template \\
\midrule

1 & \texttt{The \{adj\} \{noun1\} is resting next to the \{noun2\}.} \\

2 & \texttt{A \{noun1\} and a \{noun2\} are placed on the \{noun3\}.} \\

3 & \texttt{Please \{verb\} the \{adj\} \{noun1\} carefully.} \\

4 & \texttt{There is a \{noun1\} near the \{noun2\} and a \{noun3\} on the floor.} \\

5 & \texttt{The robot should \{verb\} the \{noun1\} without touching the \{noun2\}.} \\

6 & \texttt{A \{adj\} \{noun1\} has been spotted beside the \{noun2\}.} \\

7 & \texttt{Move the \{noun1\} towards the \{adj\} \{noun2\}.} \\

8 & \texttt{The \{noun1\} is on top of the \{noun2\} which is next to the \{noun3\}.} \\

9 & \texttt{Find the \{adj\} \{noun1\} and bring it to the \{noun2\}.} \\

10 & \texttt{The \{noun1\} fell off the \{noun2\} and landed on the \{noun3\}.} \\

\bottomrule
\end{tabular}
\end{table*}
\section{Scene--Target Evaluation Sets}
\label{app:scene-target-sets}

This appendix provides the complete scene--target sets used for the
representation analyses described in Sec.~\ref{sec:experimental-settings}.
For each scene \(s\), we define a six-element instructed-target set
\(\mathcal{T}_s\). The scene configuration and robot state are fixed across
targets in \(\mathcal{T}_s\), and only the instructed target changes.

\subsection{LIBERO-Object}
\label{app:libero-scene-targets}

All LIBERO-Object conditions use the instruction template
\begin{quote}
\small
\emph{pick up the \{target\} and place it in the basket}.
\end{quote}
For each fixed scene \(s\), the target slot is instantiated separately with
each element of \(\mathcal{T}_s\), yielding six scene--target instances.
Tables~\ref{app:libero-scene-targets} show the scenes and corresponding
target sets.

\begin{table*}[t]
    \centering
    \caption{\textbf{LIBERO-Object scene--target sets.}
    Each scene \(s\) is associated with a six-element instructed-target set
    \(\mathcal{T}_s\).}
    \label{tab:libero-scene-targets}

    \scriptsize
    \renewcommand{\arraystretch}{1.12}
    \setlength{\tabcolsep}{4pt}

    \begin{tabular}{@{}
        >{\centering\arraybackslash}m{0.05\textwidth}
        >{\centering\arraybackslash}m{0.16\textwidth}
        >{\raggedright\arraybackslash}m{0.24\textwidth}
        >{\centering\arraybackslash}m{0.05\textwidth}
        >{\centering\arraybackslash}m{0.16\textwidth}
        >{\raggedright\arraybackslash}m{0.24\textwidth}
    @{}}
        \toprule
        \textbf{} &
        \textbf{Scene} &
        \textbf{Target set} &
        \textbf{} &
        \textbf{Scene} &
        \textbf{Target set} \\
        \midrule

        1 &
        \includegraphics[width=\linewidth]
        {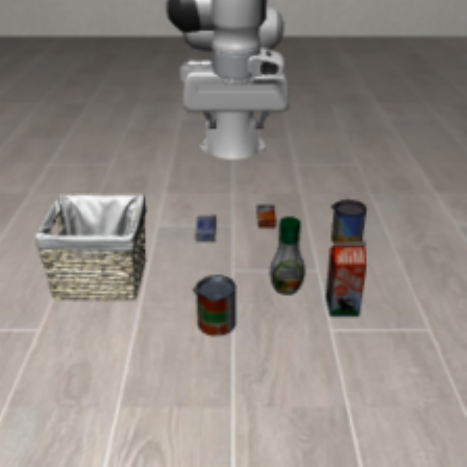}
        &
        \(\mathcal{T}_{1}=\{\)
        \textit{alphabet soup},
        \textit{salad dressing},
        \textit{cream cheese},
        \textit{milk},
        \textit{tomato sauce},
        \textit{butter}
        \(\}\)
        &
        2 &
        \includegraphics[width=\linewidth]
        {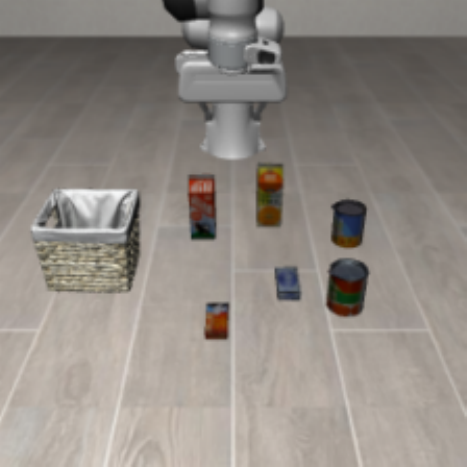}
        &
        \(\mathcal{T}_{2}=\{\)
        \textit{cream cheese},
        \textit{alphabet soup},
        \textit{milk},
        \textit{tomato sauce},
        \textit{butter},
        \textit{orange juice}
        \(\}\)
        \\[6pt]

        3 &
        \includegraphics[width=\linewidth]
        {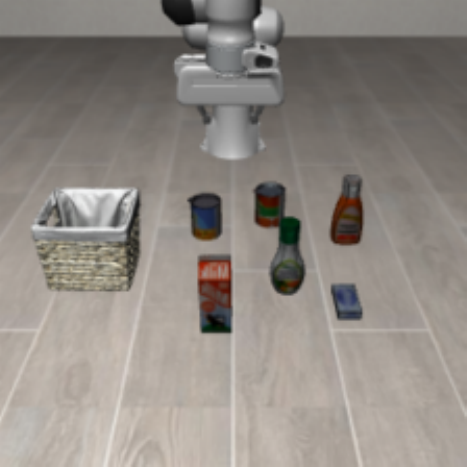}
        &
        \(\mathcal{T}_{3}=\{\)
        \textit{milk},
        \textit{cream cheese},
        \textit{tomato sauce},
        \textit{butter},
        \textit{orange juice},
        \textit{chocolate pudding}
        \(\}\)
        &
        4 &
        \includegraphics[width=\linewidth]
        {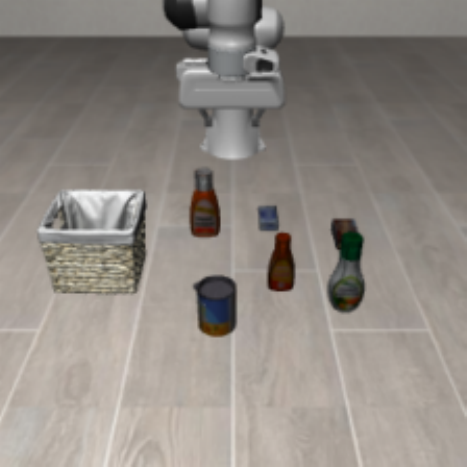}
        &
        \(\mathcal{T}_{4}=\{\)
        \textit{tomato sauce},
        \textit{milk},
        \textit{butter},
        \textit{orange juice},
        \textit{chocolate pudding},
        \textit{bbq sauce}
        \(\}\)
        \\[6pt]

        5 &
        \includegraphics[width=\linewidth]
        {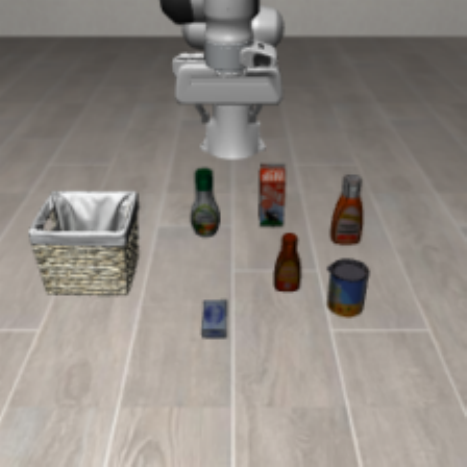}
        &
        \(\mathcal{T}_{5}=\{\)
        \textit{butter},
        \textit{tomato sauce},
        \textit{orange juice},
        \textit{chocolate pudding},
        \textit{bbq sauce},
        \textit{ketchup}
        \(\}\)
        &
        6 &
        \includegraphics[width=\linewidth]
        {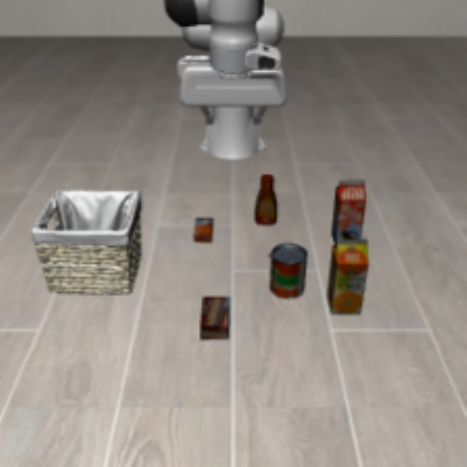}
        &
        \(\mathcal{T}_{6}=\{\)
        \textit{orange juice},
        \textit{butter},
        \textit{chocolate pudding},
        \textit{bbq sauce},
        \textit{ketchup},
        \textit{salad dressing}
        \(\}\)
        \\[6pt]

        7 &
        \includegraphics[width=\linewidth]
        {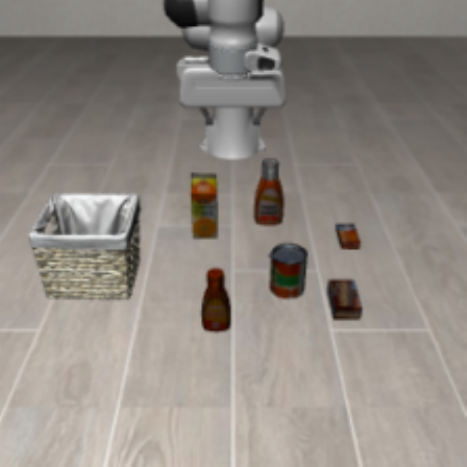}
        &
        \(\mathcal{T}_{7}=\{\)
        \textit{chocolate pudding},
        \textit{orange juice},
        \textit{bbq sauce},
        \textit{ketchup},
        \textit{salad dressing},
        \textit{alphabet soup}
        \(\}\)
        &
        8 &
        \includegraphics[width=\linewidth]
        {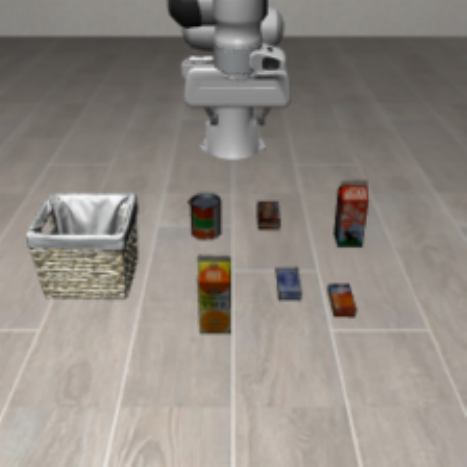}
        &
        \(\mathcal{T}_{8}=\{\)
        \textit{bbq sauce},
        \textit{chocolate pudding},
        \textit{ketchup},
        \textit{salad dressing},
        \textit{alphabet soup},
        \textit{cream cheese}
        \(\}\)
        \\[6pt]

        9 &
        \includegraphics[width=\linewidth]
        {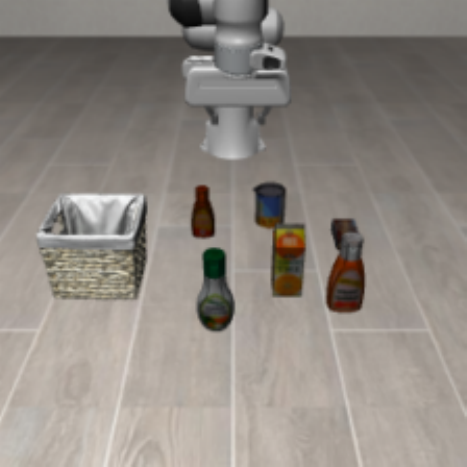}
        &
        \(\mathcal{T}_{9}=\{\)
        \textit{ketchup},
        \textit{bbq sauce},
        \textit{salad dressing},
        \textit{alphabet soup},
        \textit{cream cheese},
        \textit{milk}
        \(\}\)
        &
        10 &
        \includegraphics[width=\linewidth]
        {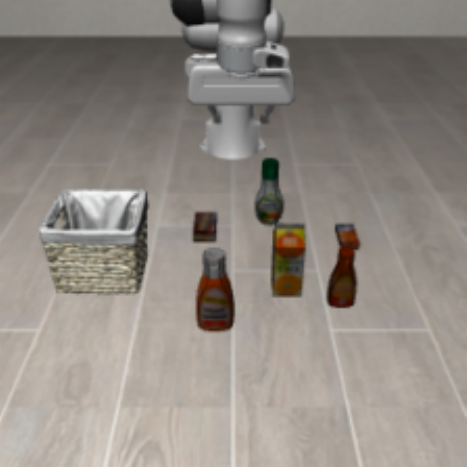}
        &
        \(\mathcal{T}_{10}=\{\)
        \textit{salad dressing},
        \textit{ketchup},
        \textit{alphabet soup},
        \textit{cream cheese},
        \textit{milk},
        \textit{tomato sauce}
        \(\}\)
        \\

        \bottomrule
    \end{tabular}
\end{table*}

\subsection{Real-World Scenes}
\label{app:real-scene-targets}

The real-world scene--target set uses the same instruction template as
LIBERO-Object,
\begin{quote}
\small
\emph{pick up the \{target\} and place it in the basket}.
\end{quote}
\begin{table*}[t]
    \centering
    \caption{\textbf{Real-world scene layouts.}
    All six scenes use the same instructed-target set
    \(\mathcal{T}_{\mathrm{real}}\) and the same instruction template.}
    \label{tab:real-scene-layouts}

    \footnotesize
    \renewcommand{\arraystretch}{1.10}
    \setlength{\tabcolsep}{4pt}

    \begin{tabular}{@{}
        >{\centering\arraybackslash}m{0.05\textwidth}
        >{\centering\arraybackslash}m{0.41\textwidth}
        >{\centering\arraybackslash}m{0.05\textwidth}
        >{\centering\arraybackslash}m{0.41\textwidth}
    @{}}
        \toprule

        1 &
        \includegraphics[width=\linewidth]
        {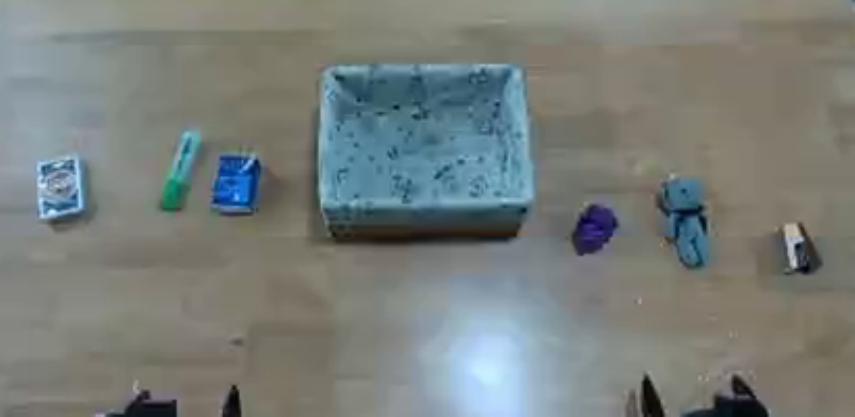}
        &
        2 &
        \includegraphics[width=\linewidth]
        {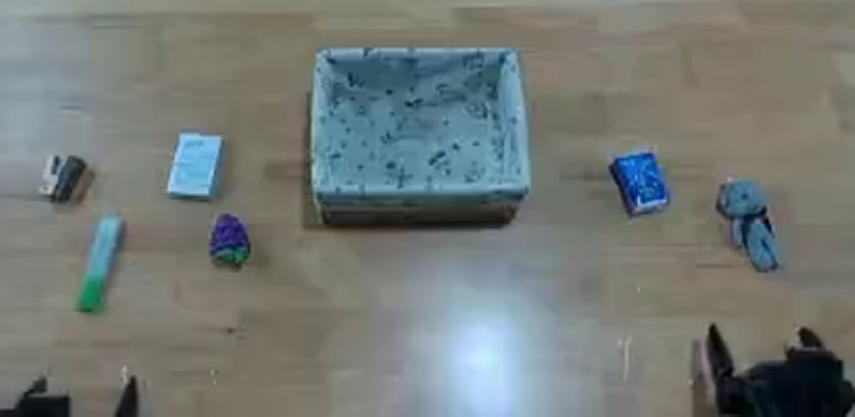}
        \\[6pt]

        3 &
        \includegraphics[width=\linewidth]
        {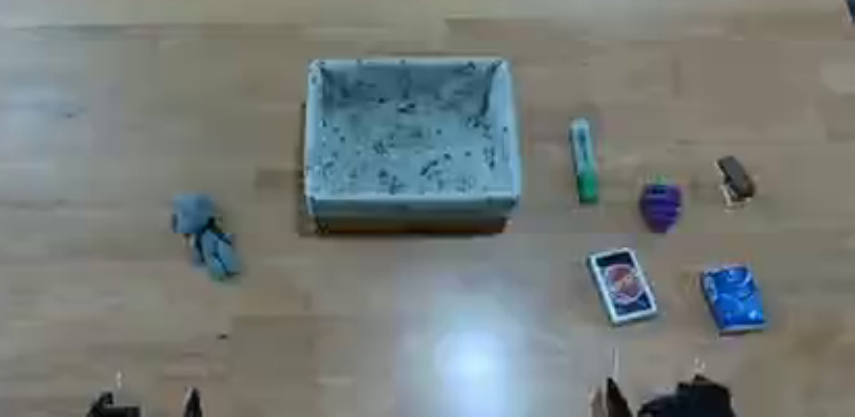}
        &
        4 &
        \includegraphics[width=\linewidth]
        {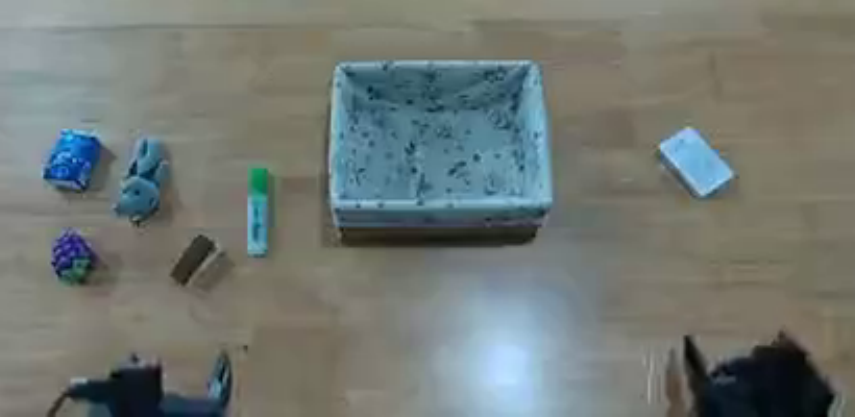}
        \\[6pt]

        5 &
        \includegraphics[width=\linewidth]
        {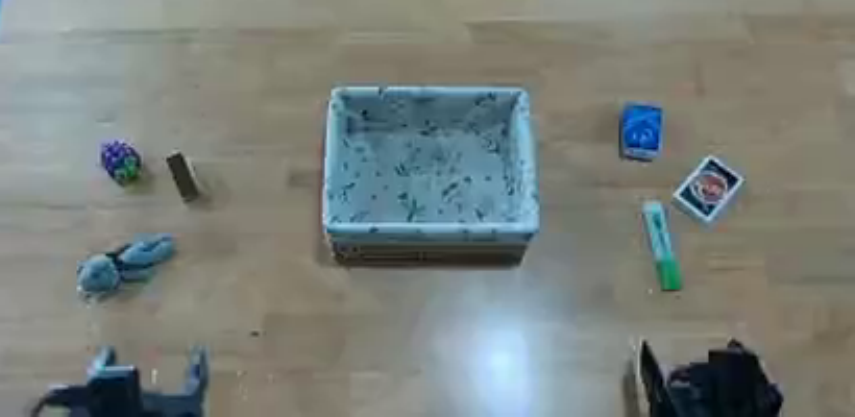}
        &
        6 &
        \includegraphics[width=\linewidth]
        {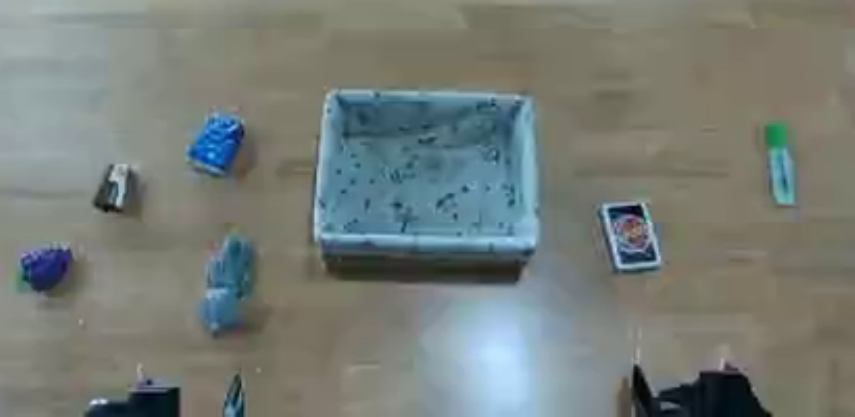}
        \\

        \bottomrule
    \end{tabular}
\end{table*}
All six scenes share the same six-element instructed-target set
\(\mathcal{T}_{\mathrm{real}}\). For each fixed scene, the target slot is
instantiated separately with each element of
\(\mathcal{T}_{\mathrm{real}}\), while only the scene layout differs across
scenes. The six scenes are shown in
Table~\ref{tab:real-scene-layouts}.

\section{Linear Probe Settings}
\label{app:linear_probe}

We use linear probes to test whether the currently instructed target and
the underlying scene identity remain explicitly accessible in the internal
instruction representations of the policy. We conduct this analysis on
$\pi_{0.5}$ and GR00T N1.7, comparing their base and fine-tuned checkpoints
on the matched scene--target evaluation sets introduced in
Sec.~\ref{sec:experimental-settings}. Policy parameters remain frozen
throughout all probing experiments.

\subsection{Probe Tasks and Evaluation Sets}

We construct two separate classification tasks.

\paragraph{Target classification.}
The target probe predicts the identity of the object specified by the
\emph{current} instruction. Importantly, the label follows the modified
instruction rather than the original task associated with the scene. For
example, when
\begin{quote}
\small
\emph{pick up the ketchup and place it in the basket}
\end{quote}
is changed to
\begin{quote}
\small
\emph{pick up the milk and place it in the basket},
\end{quote}
the target label changes from \emph{ketchup} to \emph{milk}.

\paragraph{Scene classification.}
The scene probe predicts the identity of the underlying visual scene.
Because target replacement leaves the initial environment unchanged, all
instruction variants constructed from the same initial scene share the same
scene label.

We use the matched scene--target sets described in
Sec.~\ref{sec:experimental-settings}. LIBERO-Object contains ten scenes,
each paired with six instructed targets, yielding
\[
N_{\mathrm{LIBERO}} = 10 \times 6 = 60
\]
scene--target conditions. The real-world set contains six scenes, each
paired with six instructed targets, yielding
\[
N_{\mathrm{real}} = 6 \times 6 = 36.
\]
Within each scene, the visual configuration and initial robot state remain
fixed while only the instructed target changes.

\subsection{Feature Extraction}

For each model checkpoint, we run a forward pass for every scene--target
condition and extract the hidden states associated with the instruction
tokens at each analyzed transformer layer.

Let
\[
H^{(\ell)}
=
\left[
h^{(\ell)}_1,\ldots,h^{(\ell)}_{|T|}
\right]
\]
denote the hidden states at layer $\ell$ for the lexical instruction tokens
indexed by $T$. Padding tokens and tokens outside the instruction span are
excluded. We obtain one representation per condition and layer by
mean-pooling over the instruction tokens:
\begin{equation}
\bar{h}^{(\ell)}
=
\frac{1}{|T|}
\sum_{i\in T} h^{(\ell)}_i.
\label{eq:probe_mean_pool}
\end{equation}

This produces a feature vector
\[
\bar{h}^{(\ell)} \in \mathbb{R}^{D}
\]
for every scene--target condition, where $D$ is the hidden dimension of the
corresponding model.

We perform feature-wise z-score standardization before fitting the probe.
For feature dimension $d$,
\begin{equation}
\tilde{h}^{(\ell)}_{n,d}
=
\frac{
\bar{h}^{(\ell)}_{n,d}-\mu^{(\ell)}_d
}{
\sigma^{(\ell)}_d+\epsilon
},
\label{eq:probe_zscore}
\end{equation}
where $\mu^{(\ell)}_d$ and $\sigma^{(\ell)}_d$ denote the mean and standard
deviation of feature $d$, respectively. The resulting standardized
representations are used as inputs to the linear probes.

\subsection{Label Construction}

Target identity and scene identity are treated as two independent
multiclass classification problems. For either task, a categorical label
$c_n$ is represented using a one-hot vector
\[
y_n \in \{0,1\}^{C},
\]
where $C$ is the number of classes for the corresponding prediction task
and
\[
y_{n,c}
=
\begin{cases}
1, & c=c_n,\\
0, & \text{otherwise}.
\end{cases}
\]

For target classification, $c_n$ denotes the object specified by the
current instruction. For scene classification, $c_n$ denotes the scene from
which the condition was constructed. The two probe tasks therefore use the
same representations but different label matrices.

\subsection{Ridge-Regression Probe}

For every analyzed layer, model checkpoint, environment, and prediction
task, we train an independent linear probe. Let
\[
X\in\mathbb{R}^{N\times D}
\]
contain the standardized instruction representations and
\[
Y\in\{0,1\}^{N\times C}
\]
the corresponding one-hot labels.

We use squared-loss ridge regression:
\begin{equation}
\min_{W,b}
\;
\left\|
XW+\mathbf{1}b^{\top}-Y
\right\|_F^2
+
\lambda\|W\|_F^2,
\label{eq:probe_ridge}
\end{equation}
where
\[
W\in\mathbb{R}^{D\times C}
\]
is the linear readout, $b\in\mathbb{R}^{C}$ is the intercept, and
$\lambda$ controls $L_2$ regularization.

At inference time, the predicted class is obtained from the largest
regression score:
\begin{equation}
\hat{c}_n
=
\arg\max_c
\left(
x_n^{\top}W+b
\right)_c.
\label{eq:probe_prediction}
\end{equation}

We use the same probe formulation for target and scene classification; only
the labels differ.

\subsection{Regularization Selection and Evaluation}

The regularization coefficient $\lambda$ is selected independently for each
layer and classification task using 5-fold cross-validation on the training
set. We search over logarithmically spaced values between
\[
10^{-3}
\quad\text{and}\quad
10.
\]

After selecting $\lambda$, the ridge probe is fitted using the training
samples and evaluated on held-out samples that are not used to fit the
probe. Performance is reported as multiclass classification accuracy:
\begin{equation}
\mathrm{Acc}
=
\frac{1}{N_{\mathrm{test}}}
\sum_{n=1}^{N_{\mathrm{test}}}
\mathbb{I}
\left[
\hat{c}_n=c_n
\right].
\end{equation}

A separate probe is trained for every model, checkpoint
(Base or Fine-tuned), environment (LIBERO-Object or real world), layer, and
prediction target (Object or Scene). Policy parameters are never updated by
the probing objective.

This setup measures \emph{linear accessibility}: high probe accuracy shows
that target or scene identity can be recovered through a linear readout of
the representation. It does not establish that the decoded information is
causally used by the policy or that the instructed target is correctly
grounded in the visual observation.
\section{Layer-wise Action Lens Analysis}
\label{app:logit_lens}

This section provides additional implementation details for the layer-wise
action lens analysis in Sec.~\ref{sec:action_lens}. We describe the
counterfactual replay protocol, the construction of intermediate action
estimates, the model-specific readouts, and the paired MSE used for comparison.

\subsection{Action-Lens Evaluation Protocol}

For each model, we use one successful closed-loop rollout on a
\textsc{LIBERO-Object} task under the original instruction and record the
policy inputs at every replanning event. At each recorded state, we rerun
action generation under the original instruction and the interventions defined
in Sec.~\ref{sec:experimental-settings}. For the action-lens analysis,
\textsc{Noun Replacement} includes both an out-of-scene noun and another
visible object. Each intervention instruction is fixed across all replanning
events in the rollout.

The counterfactual actions are not executed. At each replanning event, all
non-linguistic inputs are held fixed across instruction conditions, including
the observations, robot state, history or cache when applicable, denoising
schedule, and generation noise. Thus, the analysis compares instruction
conditions at matched policy states rather than through separate closed-loop
rollouts.

\subsection{Intermediate Action Readout}
Let
\begin{equation}
    \mathbf{h}_{m,r,q}^{(s,l)}
\end{equation}
denote the action-token hidden state of model $m$ at replanning event $r$,
under instruction condition $q$, after layer $l$ of denoising step $s$.
Rather than training an additional probe, we reuse the model's frozen native
action readout to decode each intermediate representation.

We denote this model-specific readout by
$\mathcal{R}_{m,s}(\cdot)$ and define the corresponding intermediate
action-space vector-field prediction as
\begin{equation}
    \widehat{\mathbf{V}}_{m,r,q}^{(s,l)}
    =
    \mathcal{R}_{m,s}
    \left(
        \mathbf{h}_{m,r,q}^{(s,l)}
    \right).
\end{equation}
Depending on the model architecture, $\mathcal{R}_{m,s}$ denotes the
model-specific composition of native output operations used to map an
intermediate action representation to the action-space vector field.
The specific instantiation of $\mathcal{R}_{m,s}$ for each model is summarized
in Table~\ref{tab:action_lens_config}.

\begin{table*}[t]
    \centering
    \small
    \setlength{\tabcolsep}{4pt}
    \renewcommand{\arraystretch}{1.08}

    \begin{tabularx}{\textwidth}{lcccX}
        \toprule
        Model
        & Denoising steps
        & Layers / step
        & Evaluated action region
        & Instantiation of $\mathcal{R}_{m,s}$ \\
        \midrule

        $\pi_{0.5}$
        & 10
        & 18
        & $10\times7$
        & Timestep-conditioned adaptive normalization followed by the native
        \texttt{action\_out\_proj} \\

        \addlinespace[3pt]

        GR00T N1.7
        & 4
        & 32
        & $16\times7$
        & Native output normalization and timestep-conditioned modulation,
        followed by the output projection and LIBERO-specific embodiment
        mapping \\

        \addlinespace[3pt]

        FastWAM
        & 10
        & 30
        & $32\times7$
        & Shared trained ActionDiT action head used in the native inference
        pathway \\

        \addlinespace[3pt]

        LingBot-VA
        & 10
        & 30
        & $16\times7$\textsuperscript{*}
        & Timestep-conditioned \texttt{norm\_out} followed by
        \texttt{action\_proj\_out} \\

        \bottomrule
    \end{tabularx}

    \caption{\textbf{Model-specific action-lens readouts.}
    $\mathcal{R}_{m,s}$ is instantiated using the native output operations of
    each model. The evaluated action region contains only the valid LIBERO
    action coordinates used in the MSE computation.
    \textsuperscript{*}For LingBot-VA, four context-fixed positions in the
    first conditioned chunk are excluded, yielding a $12\times7$ evaluated
    region for that chunk.}
    \label{tab:action_lens_config}
\end{table*}

Let $\mathbf{X}_{m,r,q}^{(s)}$ denote the noisy action state before the update
at denoising step $s$, and let $\Delta_{m,s}$ denote the signed integration
coefficient used by the model's native inference procedure. We construct the
intermediate action estimate as
\begin{equation}
    \widehat{\mathbf{A}}_{m,r,q}^{(s,l)}
    =
    \mathcal{P}_m
    \left[
        \mathbf{X}_{m,r,q}^{(s)}
        +
        \Delta_{m,s}
        \widehat{\mathbf{V}}_{m,r,q}^{(s,l)}
    \right],
\end{equation}
where $\mathcal{P}_m$ selects the valid LIBERO action horizon and the seven
action coordinates used for evaluation.

The coefficient $\Delta_{m,s}$ follows each model's native action-update rule.
All instruction conditions start from the same initial noisy action state,
\begin{equation}
    \mathbf{X}_{m,r,q}^{(0)}
    =
    \mathbf{X}_{m,r,\mathrm{orig}}^{(0)}.
\end{equation}
For $s>0$, their action states can differ because preceding denoising updates
are conditioned on different instructions. We therefore compare intermediate
readouts at matched denoising steps and network layers along each condition's
own denoising trajectory.

\subsection{MSE Computation}

For each replanning event, instruction intervention, denoising step, and
network layer, we compute the paired MSE between the intermediate action
estimate and the corresponding estimate under the original instruction:
\begin{equation}
D_{m,r,q}^{(s,l)}
=
\frac{1}{H_{m,r}C}
\sum_{h=1}^{H_{m,r}}
\sum_{c=1}^{C}
\left(
\widehat{A}_{m,r,q,h,c}^{(s,l)}
-
\widehat{A}_{m,r,\mathrm{orig},h,c}^{(s,l)}
\right)^2,
\end{equation}
where $H_{m,r}$ is the valid action horizon and $C=7$ is the number of
evaluated LIBERO action coordinates.

All distances are computed directly in each model's normalized action space;
no denormalization to physical action units is applied. Only valid LIBERO
action entries are included in the computation.

For Fig.~\ref{fig:action_lens}, we average the paired distances over all
replanning events in the selected successful rollout:
\begin{equation}
\overline{D}_{m,q}^{(s,l)}
=
\frac{1}{R_m}
\sum_{r=1}^{R_m}
D_{m,r,q}^{(s,l)},
\end{equation}
where $R_m$ denotes the number of replanning events in the analyzed rollout.
\section{UMAP Representation Analysis}
\label{app:umap_details}

This section provides the implementation details for the UMAP analysis
in Sec.~\ref{sec:umap}, including representation extraction,
normalization, and dimensionality-reduction settings.

\subsection{Evaluation Set and Representation Extraction}

The UMAP analysis uses the LIBERO-Object scene--target evaluation set
introduced in Sec.~\ref{sec:experimental-settings}. It contains ten
scenes, each paired with six instructed targets: the original target
and five alternative visible and reachable objects in the same scene,
yielding
\begin{equation}
N = 10 \times 6 = 60
\end{equation}
scene--target conditions. Within each scene, the visual configuration
and initial robot state remain fixed while only the instructed target
changes.

For both the Base and fine-tuned checkpoints of $\pi_{0.5}$, we run a
forward pass for every scene--target condition and extract the hidden
states associated with the lexical instruction tokens at each analyzed
transformer layer. Padding tokens and tokens outside the instruction
span are excluded.

Let
\begin{equation}
H_n^{(\ell)}
=
\left[
h_{n,1}^{(\ell)},\ldots,h_{n,|T_n|}^{(\ell)}
\right]
\end{equation}
denote the hidden states at layer $\ell$ for the lexical instruction
tokens indexed by $T_n$ in condition $n$. We obtain one representation
per condition and layer by mean-pooling over these tokens:
\begin{equation}
\bar{h}_n^{(\ell)}
=
\frac{1}{|T_n|}
\sum_{i\in T_n}
h_{n,i}^{(\ell)}.
\label{eq:umap_mean_pool}
\end{equation}

This produces one feature vector
$\bar{h}_n^{(\ell)}\in\mathbb{R}^{D}$
for each scene--target condition and analyzed layer, where $D$ is the
hidden dimension. The representation-extraction procedure follows the
same instruction-token pooling protocol used in the linear probe
analysis in App.~\ref{app:linear_probe}.

\subsection{Feature Normalization}

Before dimensionality reduction, we apply feature-wise z-score
standardization across the 60 LIBERO-Object scene--target conditions.
For each checkpoint and analyzed layer, let
\begin{equation}
X^{(\ell)}
=
\begin{bmatrix}
(\bar{h}_1^{(\ell)})^\top\\
\vdots\\
(\bar{h}_N^{(\ell)})^\top
\end{bmatrix}
\in\mathbb{R}^{N\times D}.
\end{equation}

For feature dimension $d$, the standardized representation is
\begin{equation}
\tilde{h}_{n,d}^{(\ell)}
=
\frac{
\bar{h}_{n,d}^{(\ell)}-\mu_d^{(\ell)}
}{
\sigma_d^{(\ell)}+\epsilon
},
\label{eq:umap_zscore}
\end{equation}
where $\mu_d^{(\ell)}$ and $\sigma_d^{(\ell)}$ denote the mean and
standard deviation of feature $d$ across the 60 conditions,
respectively. Standardization is performed independently for each
checkpoint and analyzed layer.

\subsection{UMAP Dimensionality Reduction}

For each checkpoint and analyzed layer, we independently apply UMAP to
the standardized representation matrix
$\tilde{X}^{(\ell)}\in\mathbb{R}^{60\times D}$ and obtain a
two-dimensional embedding:
\begin{equation}
Z^{(\ell)}
=
\operatorname{UMAP}\!\left(\tilde{X}^{(\ell)}\right)
\in\mathbb{R}^{60\times 2}.
\label{eq:umap_projection}
\end{equation}

The Base and fine-tuned checkpoints are embedded independently at each
layer. We therefore interpret the visualization through the
within-panel clustering structure rather than through absolute
coordinates or distances across panels.

We use the same UMAP settings for all checkpoint--layer pairs:
\texttt{n\_neighbors}=10,
\texttt{min\_dist}=0.15, and
\texttt{metric}=\texttt{euclidean}.
All embeddings use two output dimensions.

Figure~\ref{fig:umap_all} reports representative layers 1, 6, 13, and
18. Scene identity is encoded by color and instructed-target identity
by marker shape. These labels are used only for visualization and are
not provided to UMAP during dimensionality reduction.
\section{Shared Semi-NMF and Variance Partitioning}
\label{app:semi_nmf_details}

\paragraph{Shared semi-NMF.}
We adapt the semi-NMF implementation in NMFLibrary (\url{https://github.com/hiroyuki-kasai/NMFLibrary}) to jointly factorize matched base and fine-tuned instruction representations. At each analyzed layer, we stack the two representation matrices:
\begin{equation}
X_{\mathrm{joint}}
=
\begin{bmatrix}
X_{\mathrm{base}}\
X_{\mathrm{FT}}
\end{bmatrix}
\in\mathbb{R}^{2N\times D},
\end{equation}
where $N=60$ for LIBERO-Object. We fit $K=12$ components by solving
\begin{equation}
\min_{A\geq 0,;B}
\frac{1}{2}\left|X_{\mathrm{joint}}-AB^\top\right|_F^2,
\label{eq:app_semi_nmf}
\end{equation}
where $A\in\mathbb{R}_{\geq 0}^{2N\times K}$ contains nonnegative activations and $B\in\mathbb{R}^{D\times K}$ is an unconstrained signed basis. The first and last $N$ rows of $A$ correspond to the base and fine-tuned checkpoints, respectively. The shared basis therefore pairs components across checkpoints within each layer.

\paragraph{Variance partitioning.}
For each component and checkpoint, we regress activations across observed conditions on categorical scene labels, target labels, and both labels jointly. All three models include an intercept; the joint model contains additive main effects without interactions. Let their ordinary coefficients of determination be $R_S^2$, $R_T^2$, and $R_{ST}^2$. We partition activation variance as
\begin{equation}
\begin{aligned}
U_{\mathrm{scene}} &= R_{ST}^2-R_T^2,\
U_{\mathrm{target}} &= R_{ST}^2-R_S^2,\
C_{\mathrm{shared}} &= R_S^2+R_T^2-R_{ST}^2,\
R_{\mathrm{res}} &= 1-R_{ST}^2.
\end{aligned}
\label{eq:app_variance_partition}
\end{equation}
The unique terms measure the additional variance explained by each factor after controlling for the other. The shared term captures overlapping explanatory variation and can be negative under suppression; the residual is unexplained by the additive model.

We plot each component at $(U_{\mathrm{target}},U_{\mathrm{scene}})$ and connect its base and fine-tuned positions. An upper-left shift indicates decreased target-unique variance and increased scene-unique variance. These quantities describe representational organization, not causal contributions to behavior.

\section{Detailed Literature Review}
\label{app:detailed_literature_review}
Recent advances in robot learning have expanded the ways perception is translated into action. This section organizes the literature into four complementary paradigms. \textbf{Vision–Action (VA) models} learn direct mappings from visual observations to motor commands, with an emphasis on precise sensorimotor control. \textbf{Vision–Language–Action (VLA)} models condition behavior on natural-language instructions, enabling a single policy to address a broader range of tasks. \textbf{World–Action Models (WAMs)} couple action generation with predictions of how the environment will evolve, providing a basis for anticipating action outcomes.

\subsection{Vision--Action Models.}
Vision--action (VA) models learn visuomotor policies that map visual observations and robot states to actions without requiring language instructions. Their development has centered on three related challenges: representing spatial information, generating coherent action sequences, and maintaining robust control during execution.

Diffusion-based policies have become a prominent approach to modeling multimodal action distributions. Diffusion Policy~\citep{chi2025diffusion} established action-sequence denoising as an effective framework for visuomotor control. Subsequent work has strengthened its spatial representations: DP3~\citep{ze20243d} conditions action generation on 3D point clouds, while iDP3~\citep{ze2025generalizable} extends this direction to humanoid manipulation learned from human demonstrations. Other methods refine how visual information enters the policy. H$^3$DP~\citep{lu2026h} introduces hierarchies across observations, visual features, and action generation; 3D Diffuser Actor~\citep{ke20243d} uses tokenized 3D scene representations; and Render and Diffuse (R\&D)~\citep{vosylius2024render} aligns observations and candidate actions by rendering robot configurations in image space. Mamba Policy~\citep{cao2025mamba} explores a more efficient sequence-modeling backbone, whereas MCDP~\citep{cao2025modality} combines diffusion policies trained on different visual modalities at inference time. GCP~\citep{cao2026compose} further studies distribution-level policy composition across diffusion- and flow-based policies.

A second line of work addresses the structure and reliability of generated actions. DexDiffuser~\citep{weng2024dexdiffuser} targets dexterous grasp generation, while ALOHA Unleashed~\citep{zhao2024aloha} applies diffusion-based action generation to complex bimanual manipulation. ChainedDiffuser~\citep{xian2023chaineddiffuser} connects end-effector keyposes to executable trajectories, and HDP~\citep{ma2024hierarchical} incorporates kinematic structure into policy learning. For generalization across objects and scenes, S$^2$-Diffusion~\citep{yang2025s} investigates the transfer from instance-level demonstrations to category-level skills, and C3DM~\citep{saxena2023constrained} uses constrained visual context to reduce the influence of distractors. Crossway Diffusion~\citep{li2024crossway} improves learned representations through an auxiliary self-supervised objective, while Equivariant Diffusion Policy~\citep{wang2024equivariant} builds geometric symmetry into action generation.

Temporal consistency and responsiveness have motivated complementary advances. Streaming Diffusion Policy~\citep{hoeg2024streaming} addresses continuous action generation during execution, and Bidirectional Decoding (BID)~\citep{liu2025bidirectional} improves action-chunk execution through guided test-time sampling. Beyond diffusion, ACT~\citep{zhao2023learning} predicts action chunks with a transformer-based policy, while Chunking Causal Transformer~\citep{zhang2025autoregressive} models dependencies across chunks. Act3D~\citep{gervet2023act3d} and Spatial Policy~\citep{liu2025spatial} emphasize spatially grounded action prediction. Flow matching provides another route to generative control: ManiFlow~\citep{yan2025maniflow} and VITA~\citep{gao2026vita} develop flow-based visuomotor policies, while Flow Matching Policy Gradients~\citep{mcallister2026flow} connects flow-based policies with reinforcement learning. Finally, DynaGuide~\citep{du2026dynaguide} uses feedback from a dynamics model to steer a diffusion policy at inference time, and Latent Policy Barrier~\citep{sun2026latent} uses learned latent boundaries to identify and respond to states outside the expert distribution.

\subsection{Vision-Language-Action Models (VLA)}

Vision-language-action (VLA) models extend visuomotor control by conditioning robot behavior on natural-language instructions. Their central challenge is to connect semantic information---such as the requested object, goal, or task---to actions that are executable in the observed scene. Existing approaches differ in where this connection is made: some learn language-conditioned policies directly, while others adapt pretrained vision-language models for action prediction or use language models to guide a separate control policy.

Early language-conditioned approaches explored several ways to ground goals in behavior. MDT~\citep{reuss2024multimodal} learns a diffusion policy from language and visual goal specifications, while PlayFusion~\citep{chen2023playfusion} learns skills from language-annotated play data. StructDiffusion~\citep{liu2022structdiffusion} uses language and object-centric 3D observations to generate physically plausible rearrangement goals. Rather than changing the policy architecture, ROSIE~\citep{yu2023scaling} uses text-guided image generation to augment robot training data. These works establish complementary routes from linguistic goals to visuomotor behavior.

Scaling beyond individual tasks and platforms has motivated training on diverse robot data. RT-1~\citep{brohan2023rt} demonstrated large-scale instruction-conditioned robot learning, while RT-2~\citep{brohan2023rt} transferred knowledge from web-trained vision-language models into robotic action prediction. The Open X-Embodiment effort and RT-X models~\citep{o2024open} extended this direction across datasets and embodiments; Octo~\citep{team2024octo} likewise trains a generalist policy on heterogeneous robot demonstrations. Other methods address the scarcity or incompatibility of action data. LAPA~\citep{ye2025latent} learns latent action representations from videos without requiring ground-truth robot actions during pretraining. RDT-1B~\citep{liu2025rdt} uses a diffusion transformer and a unified action representation for bimanual manipulation, while RDT2~\citep{liu2026rdt2} investigates cross-embodiment generalization through larger-scale UMI data. TinyVLA~\citep{wen2025tinyvla} instead emphasizes a compact vision-language backbone and a diffusion action decoder for efficient adaptation.

A major architectural question is how to preserve semantic reasoning while generating precise continuous actions. DiffusionVLA~\citep{wen2025diffusionvla} and HybridVLA~\citep{liu2025hybridvla} combine autoregressive modeling with diffusion-based action generation. The $\pi_0$ family~\citep{black2024pi_0,intelligence2025pi_,intelligence2026pi} pairs a pretrained vision-language backbone with a flow-based action expert, with successive models extending its generalization and conditioning capabilities. ChatVLA~\citep{zhou2025chatvla} addresses the interaction between multimodal understanding and robot control through phased training and an expert-based architecture. Together, these models illustrate different ways to connect representations learned from vision and language to action generation.

Spatial reasoning and longer-horizon planning form a related direction. 3D-VLA~\citep{zhen20243d} connects 3D perception and language reasoning with generative predictions of future scenes, providing a bridge to world-action modeling. SpatialBot~\citep{cai2025spatialbot} strengthens depth-aware spatial understanding in vision-language models. At the planning level, HiP~\citep{ajay2023compositional} composes models specialized for language, visual prediction, and control, while Plan Diffuser~\citep{sharan2024plan} grounds language-model plans through diffusion-based generation. ReKep~\citep{huang2024rekep} expresses tasks as spatial and temporal keypoint constraints, whereas VoxPoser~\citep{huang2023voxposer} converts language-guided affordances into 3D value maps. These planning systems connect semantic goals to robot execution and lead into the agent-based approaches discussed later.

\subsection{World-Action Models (WAM)}

World-action models (WAMs) connect robot control with predictions of how the environment may change. This predictive signal can play different roles: a model may generate a visual plan before choosing actions, learn future observations and actions jointly, or use future prediction only as supervision during training. The distinction matters because forecasting a useful future representation does not necessarily require rendering future video at deployment. UniPi~\citep{du2023learning} provides an early example of planning through text-conditioned video generation, treating predicted visual trajectories as an intermediate representation for decision-making.

One approach separates visual prediction from action execution. Gen2Act~\citep{bharadhwaj2024gen2act} generates videos of human behavior in novel scenes and uses them to guide robot manipulation. Vidar~\citep{feng2025vidar} combines video diffusion pretraining with a masked inverse dynamics model that recovers actions, while CoT-VLA~\citep{zhao2025cot} autoregressively predicts future images as visual goals before generating action sequences. Act2Goal~\citep{zhou2025act2goal} connects a goal-conditioned visual world model to control at multiple temporal scales. Related work uses video models to expand training data or support visual planning: DreamGen~\citep{jang2025dreamgen} produces synthetic robot trajectories from generated videos, and RoboDreamer~\citep{zhou2024robodreamer} factorizes video generation to compose visual plans from learned elements.

A second approach learns visual dynamics and actions within a shared model. GR-1~\citep{wu2024unleashing} predicts future images alongside robot actions, and GR-2~\citep{cheang2024gr} extends video pretraining to web-scale data before adapting to robot trajectories. WorldVLA~\citep{cen2025worldvla} and RynnVLA-002~\citep{cen2025rynnvla} further unify action prediction with world modeling. LingBot-VA~\citep{li2026causal} learns frame prediction and policy execution within an autoregressive diffusion framework. Unified World Models (UWM)~\citep{zhu2025unified} couple video and action diffusion in a model that can be used for policy learning, visual prediction, and dynamics modeling. DreamVLA~\citep{zhang2026dreamvla} instead predicts compact world knowledge to support inverse dynamics and action generation.

Large video models offer another source of predictive representations. Cosmos Policy~\citep{kim2026cosmos} adapts a pretrained video foundation model to visuomotor control through post-training, while DreamZero~\citep{ye2026world} builds a joint video-action model on a pretrained video diffusion backbone. Genie Envisioner~\citep{liao2026genie} brings policy learning, visual generation, and evaluation into a video-generative framework. DreamDojo~\citep{gao2026dreamdojo} pretrains an action-controllable world model on large-scale egocentric human video before adapting it to robot data. At the level of model design, DUST~\citep{won2025dual} uses separate but interacting diffusion streams for visual futures and actions. Motus~\citep{bi2026motus} and MotuBrain~\citep{team2026motubrain} use multiple transformer streams to model understanding, video generation, and action within shared systems. LDA-1B~\citep{lyu2026lda} scales joint learning of dynamics, policy, and visual forecasting across heterogeneous embodied data.

Explicit future-image generation, however, adds computational cost and may place demands on a model that differ from those of action prediction. FLARE~\citep{zheng2025flare} introduces future-latent prediction as an implicit world-modeling objective for policy learning. Fast-WAM~\citep{yuan2026fast} retains video co-training but omits future generation at test time, allowing a direct comparison between the benefits of predictive training and inference-time imagination. GigaWorld-Policy~\citep{ye2026gigaworld} likewise centers action prediction while using visual dynamics to inform training. Across these methods, the central design choice is not simply whether to predict the future, but \emph{what future information to learn, how to connect it to actions, and whether it must be generated during execution}.

\subsection{Agentic Robotics}

Agentic robotics studies how a robot can pursue a task through repeated
observation, reasoning, action, and feedback. In these systems, a
multimodal model may decompose a goal, select executable skills, inspect
their outcomes, and revise its plan as the scene changes. The agent
therefore coordinates perception and control over an entire task,
rather than producing a single action from the current observation.

Recent work has made this coordination more explicit. Gemini Robotics
1.5~\citep{team2025gemini, team2025gemini15} combines embodied reasoning with a
multi-embodiment action model. Reflective Planning~\citep{feng2025reflective}
uses imagined future states to reconsider skill choices, while
RoboAgent~\citep{xu2026roboagent} learns to chain basic capabilities
for embodied task planning. MoMaStage~\citep{li2026momastage} grounds
planning in a skill-state graph, and Physical Agency~\citep{galanti2026addressing}
places a closed-loop orchestrator above existing policies and robot
skills. RoboClaw~\citep{li2026roboclaw} extends agent coordination
across data collection, policy learning, and task execution.

A newer line of work asks how much of the policy itself a general-purpose
agent can provide. Agent as Policy~\citep{jia2026agent} lets an agent
inspect observations, write programs, issue robot commands, and adapt
from physical feedback. GPT-Policy~\citep{cheng2026context} uses
demonstrations and interaction history as context for verified robot
tool calls. Evaluation of GPT-6 Astra examine direct action
selection and its combination with a pretrained policy
\citep{su2026astra}; their results
also expose the remaining difficulty of precise and dynamic control. Finally, GaP~\citep{chen2026gap} uses coding agents and structured feedback to
improve robot policies or executable skill graphs. Together, these
studies shift attention toward the interfaces and feedback loops that
connect general reasoning to reliable physical execution.
\end{document}